\documentclass[final]{cvpr}

\usepackage{times}
\usepackage{epsfig}
\usepackage{graphicx}
\usepackage{amsmath}
\usepackage{amssymb}
\usepackage{balance}
\usepackage{amsmath}
\usepackage{amssymb}
\usepackage{xspace}
\usepackage{algpseudocode}
\usepackage{comment}
\usepackage{dsfont}
\usepackage{microtype}  %
\usepackage{appendix}

\DeclareMathOperator{\argmin}{argmin}

\DeclareMathOperator{\IoU}{IoU}
\DeclareMathOperator{\stopgradient}{sg}
\newcommand{\sg}[1]{\stopgradient[#1]}
\renewcommand{\vec}[1]{\boldsymbol{\mathbf{#1}}}
\newcommand{\veci}[2]{\vec{#1}_{#2}}                %
\newcommand{\set}[1]{\mathcal{#1}}                  %
\newcommand{\field}[1]{\mathbb{#1}}

\newcommand{\Lovasz}{Lov\'{a}sz\xspace}
\newcommand{\Lovaszmath}{Lov\acute{a}sz\xspace}
\newcommand{\dcos}[2]{d_\text{cos}({#1}, {#2})}
\DeclareMathOperator{\LovaszBinary}{\Lovaszmath{}Binary}
\DeclareMathOperator{\LovaszSoftmax}{\Lovaszmath{}Softmax}

\newif\ifsuppmat
\suppmatfalse
\suppmattrue %

\usepackage[pagebackref=true,breaklinks=true,colorlinks,bookmarks=false]{hyperref}

\newif\ifsuppmat
\suppmatfalse
\suppmattrue %

\def\cvprPaperID{6158} %
\def\confYear{CVPR 2021}

\begin{document}

\title{Hierarchical \Lovasz Embeddings for Proposal-free Panoptic Segmentation}

\newcommand{\authorlist}[1][]{%
Tommi Kerola\textsuperscript{1}{#1}\quad
Jie Li\textsuperscript{2}\quad
Atsushi Kanehira\textsuperscript{1}\quad
Yasunori Kudo\textsuperscript{1}\quad
Alexis Vallet\textsuperscript{1}\quad
Adrien Gaidon\textsuperscript{2}\\
\textsuperscript{1}Preferred Networks, Inc.\quad
\textsuperscript{2}Toyota Research Institute (TRI)
}
\author{
\authorlist[\thanks{Correspondence to {\tt\small tommi@preferred.jp}}]{}
}

\makeatletter
\let\newtitle\@title
\let\newauthor\@author
\makeatother

\maketitle

\begin{abstract}
Panoptic segmentation brings together two separate tasks: instance and semantic segmentation. Although they are related, unifying them faces an apparent paradox: how to learn simultaneously instance-specific and category-specific (i.e.~instance-agnostic) representations jointly.
Hence, state-of-the-art panoptic segmentation methods use complex models with a distinct stream for each task.
In contrast, we propose Hierarchical \Lovasz Embeddings, per pixel feature vectors that simultaneously encode instance- and category-level discriminative information.
We use a hierarchical \Lovasz hinge loss to learn a low-dimensional embedding space structured into a unified semantic and instance hierarchy without requiring separate network branches or object proposals.
Besides modeling instances precisely in a proposal-free manner, our Hierarchical \Lovasz Embeddings generalize to categories by using a simple Nearest-Class-Mean classifier, including for non-instance ``stuff'' classes where instance segmentation methods are not applicable.
Our simple model achieves state-of-the-art results compared to existing proposal-free panoptic segmentation methods on Cityscapes, COCO, and Mapillary Vistas. Furthermore, our model demonstrates temporal stability between video frames.
\end{abstract}

\section{Introduction}
Holistic scene understanding is an important task in computer vision, where a model is trained
to explain each pixel in an image, whether that pixel describes \emph{stuff} -- uncountable regions
of similar texture such as grass, road or sky -- or \emph{thing} -- a countable object with individually identifying characteristics, such as people or cars.
While holistic scene understanding received some early attention~\cite{tu2005image,yao2012describing,tighe2014scene},
modern deep learning-based methods have mainly tackled the tasks of modeling stuff and things independently under the task names semantic segmentation and instance segmentation.
Recently, Kirillov et al. proposed the panoptic quality (PQ) metric for unifying these two parallel
tracks into the holistic task of panoptic segmentation~\cite{kirillov2019panoptic}.
Panoptic segmentation is a key step for visual understanding, with applications in fields such as
autonomous driving or robotics, where it is crucial to know both the locations of dynamically trackable things, as well as
static stuff classes. For example, an autonomous car needs to be able to both avoid other cars with high precision, as well as understand the location of the road and sidewalk to stay on a desired path.

\begin{figure}[t]
{\centering
   \def\figwidth{0.33}%
   \includegraphics[width=\figwidth\linewidth]{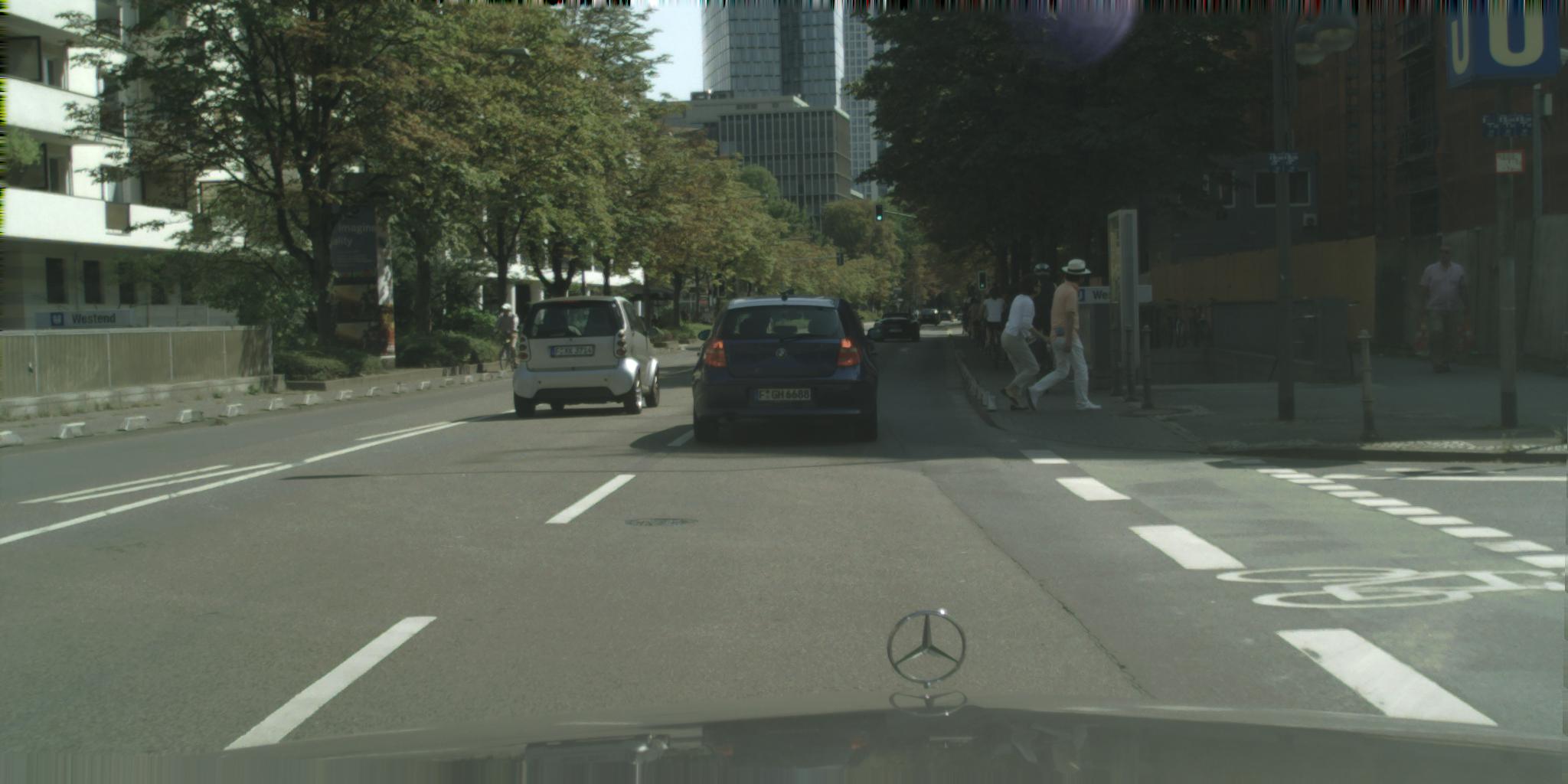}%
   \includegraphics[width=\figwidth\linewidth]{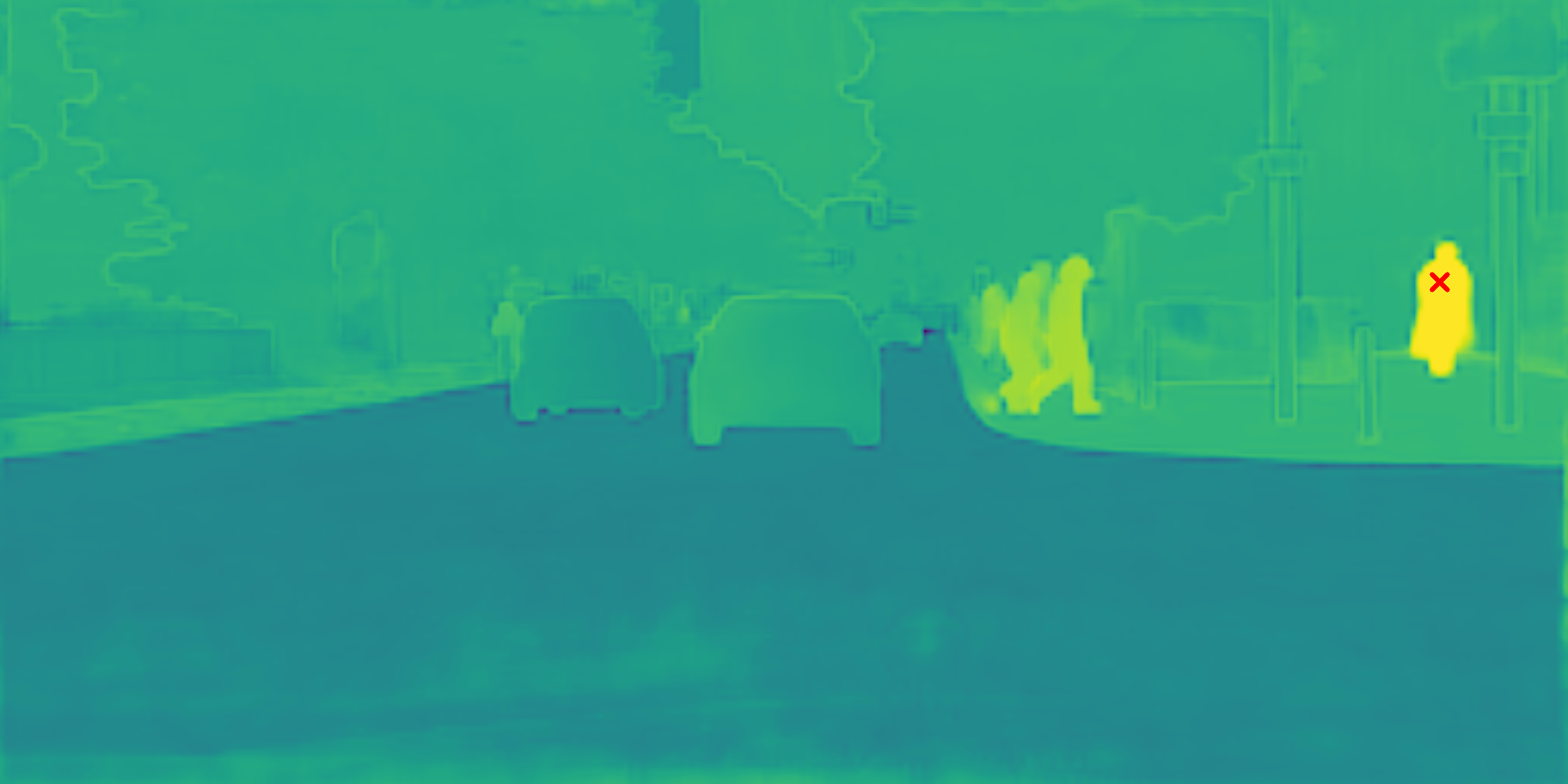}%
   \includegraphics[width=\figwidth\linewidth]{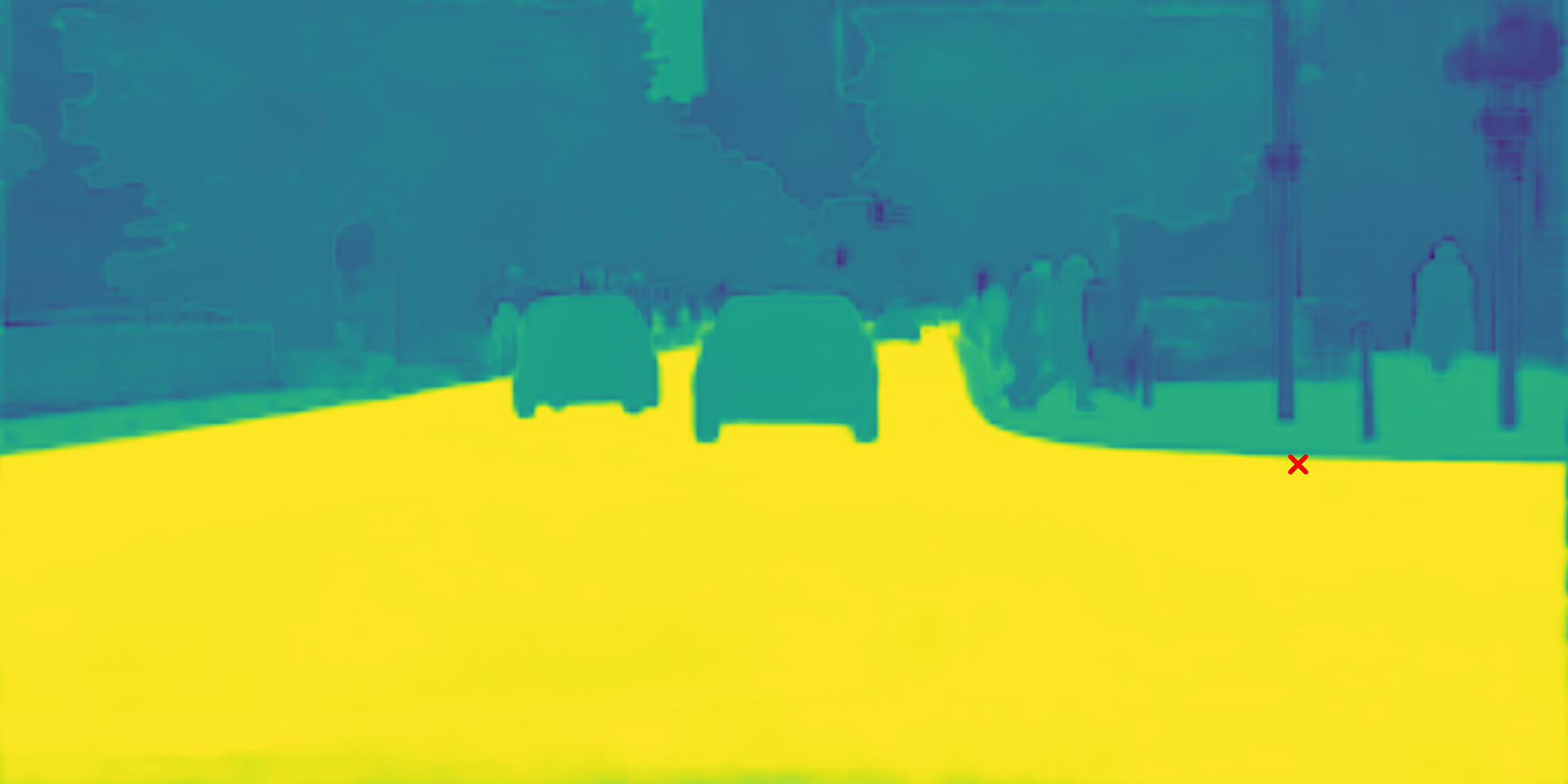}
   \includegraphics[width=\figwidth\linewidth]{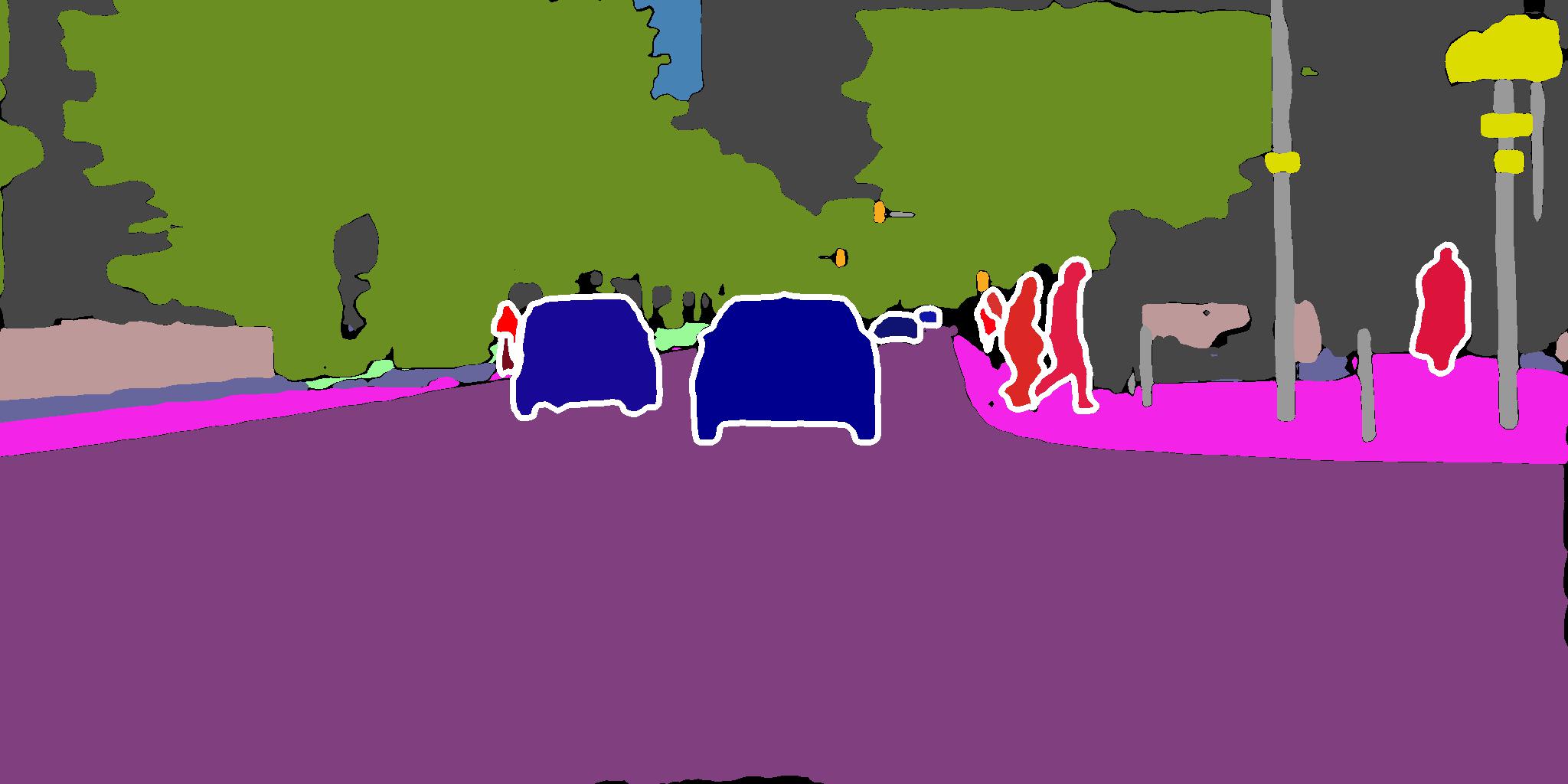}%
   \includegraphics[width=\figwidth\linewidth]{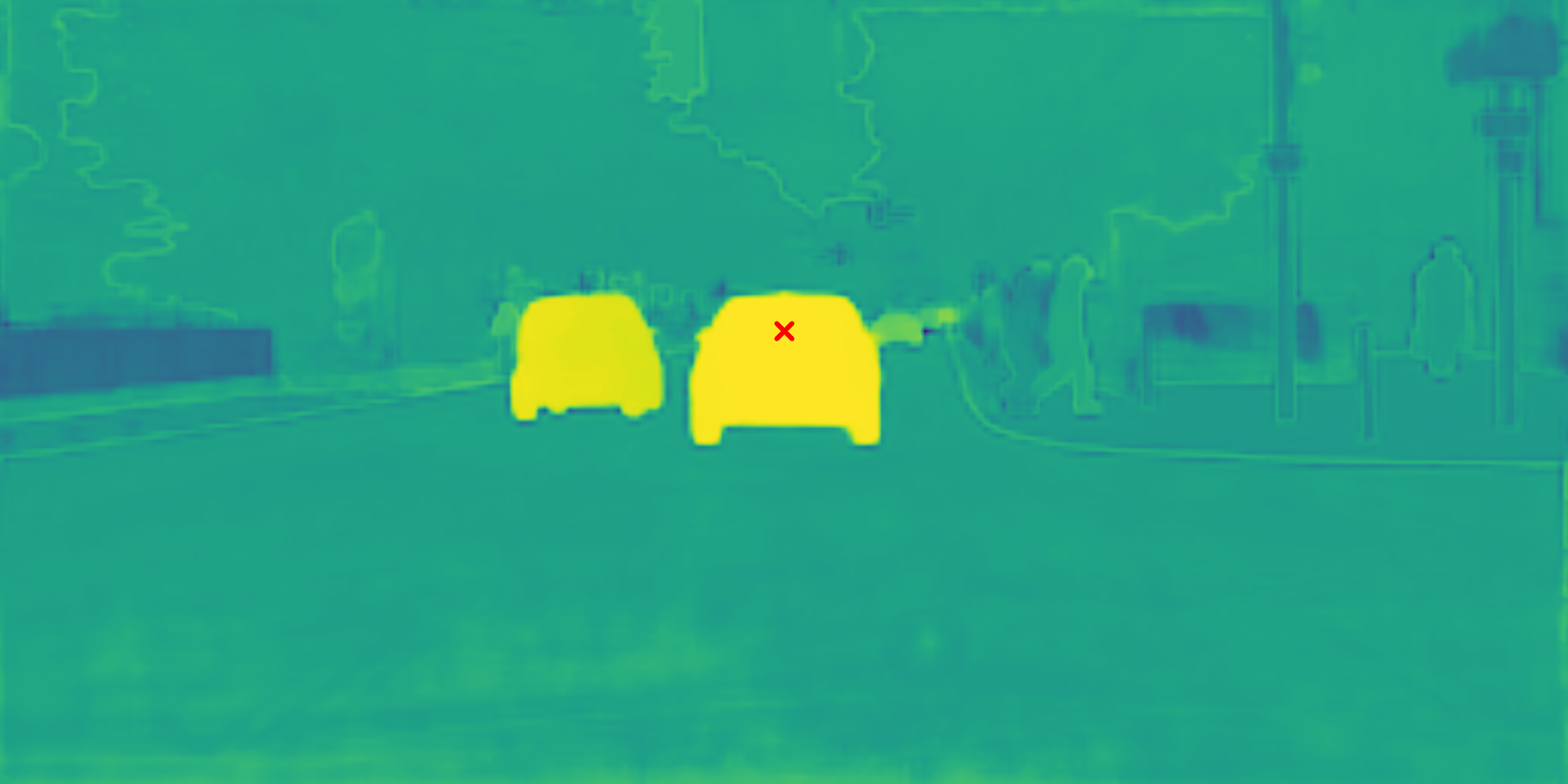}%
   \includegraphics[width=\figwidth\linewidth]{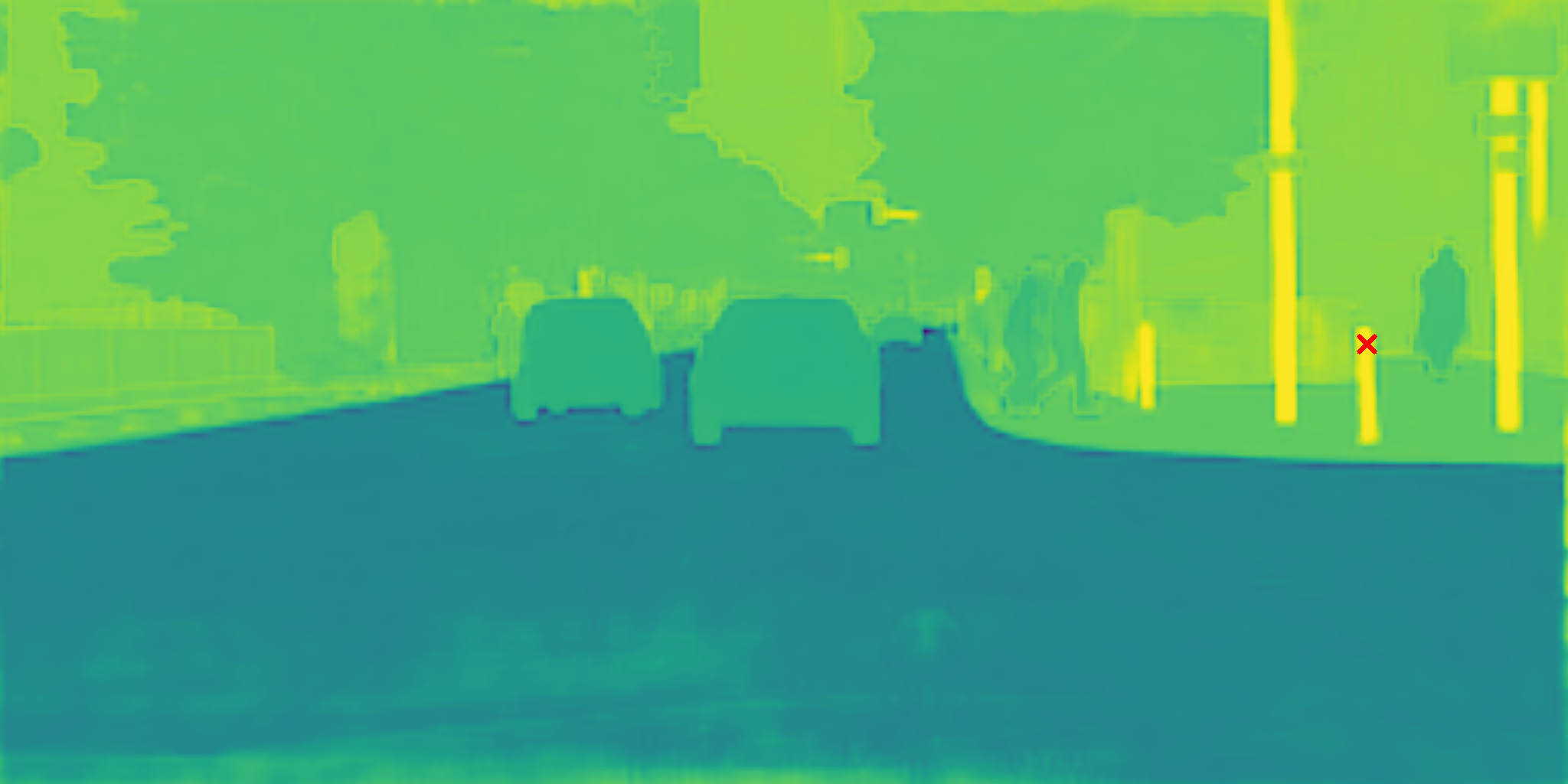}%
   \caption{Example panoptic segmentation using our method (bottom left). Predictions are decoded from our Hierarchical \Lovasz Embeddings, which encode both instance and class information. The rightmost columns illustrate distances in embedding space between all pixels and a target pixel in red (warmer colors denote smaller distances). We can see the hierarchical structure: pixels on the same instance being closest, and pixels on other instances in the same category being closer than pixels from other categories.
   }%
   \label{fig:method-abstract}%
}%
\vspace{-1.05em}
\end{figure}%

\begin{figure*}[t]
{\centering
   \includegraphics[width=1.0\linewidth]{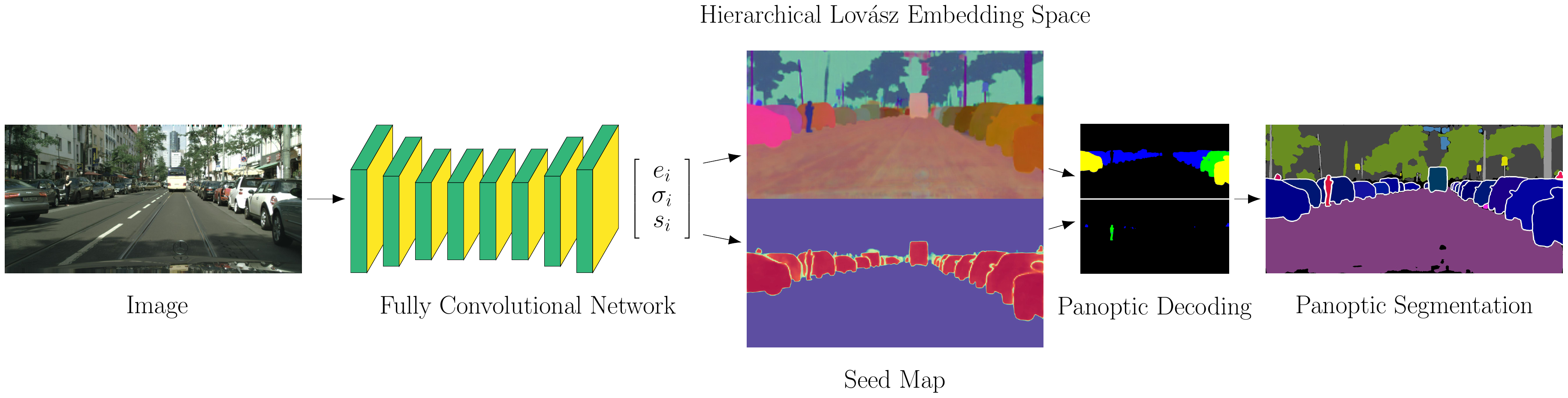}
   \vspace{-1em}
   \caption{Overview of our method. We train a single-shot fully convolutional network to predict for each pixel $i$ a hierarchical embedding $\veci{e}{i}$ as well as an instance seed $s_i$ and variance $\sigma_i$. The seed map represents probable instance locations, and the variance defines the margins of the hierarchical embedding space. These are used for panoptic decoding of the embedding space.}%
   \label{fig:method-overview}
   \vspace{-1em}
}%
\end{figure*}%

A strong but complex baseline for panoptic segmentation is to run independent methods
for semantic segmentation and instance segmentation, and then fusing the results.
To improve upon this, previous works combine both tasks in a joint model~\cite{xiong2019upsnet,kirillov2019panopticfpn,li2018learning}.
Early methods focus more on a joint model and the majority of them leverage two-stage instance detection models~\cite{porzi2019seamless,xiong2019upsnet,kirillov2019panopticfpn,li2019attention,degeus2019panoptic}; some recent works propose 
bottom-up~\cite{gao2019ssap,cheng2020panoptic,yang2019deeperlab}, yet instance and semantic segmentation are still treated separately. Performing panoptic segmentation as a single task without duplicated information across sub-tasks remains an interesting question.

Intuitively, instances are contained in semantics, where semantic representations have higher variance in the embedding space to describe a general category, whereas instances have smaller variance to capture object-specific characteristics. This constitutes a natural hierarchical relationship between instances and semantics (cf. Figure~\ref{fig:method-hierarchy}).

In this work, we propose to model panoptic segmentation as a unified task via a novel formulation of the problem: learning hierarchical pixel embeddings.
Creating a unified embedding for the task opens up the potential of leveraging embedding space analysis as conducted in the natural language processing community~\cite{mikolov2013distributed,morin2005hierarchical,mnih2009scalable,miller1995wordnet}.

Leveraging hierarchical structure for representation learning is well studied~\cite{goodman2001classes,deng2009imagenet,redmon2017yolo9000}.
However, previous work has not taken advantage of the semantic-instance visual hierarchy for end-to-end unified scene parsing.
In this paper, we leverage advances in structural representation learning and encode ``instance'' and ``category'' features in a hierarchical embedding space. By doing so, we reduce the redundant information in the output space and optimize the information efficiency of network parameters for panoptic segmentation. 

\textbf{Our main contribution} is a novel representation learning approach for panoptic segmentation that treats it as a unified task.
We propose a simple architecture and loss to learn pixel-wise embeddings to represent instances, object categories and stuff classes, thus enabling \textbf{unified embedding-based single-shot panoptic segmentation}.
In particular, we leverage the \Lovasz hinge loss to learn a structured Hierarchical \Lovasz Embedding space where categories can be represented with categories jointly.
An overview of our method is shown in Figure~\ref{fig:method-overview}.

Compared to conventional panoptic segmentation models, our model displays temporal stability between video frames, creating temporal smoothness in predictions that can be directly used in downstream applications such as object tracking and prediction in autonomous driving or mobile robotic systems, where data association is a key component (cf. Figure~\ref{fig:cityscapes-temporal-stability}).
Experiments on the Cityscapes~\cite{cordts2016cityscapes}, COCO~\cite{lin2014microsoft} and Vistas~\cite{neuhold2017mapillary} datasets show that our method establishes state-of-the-art results for proposal-free methods, and also yields competitive results compared with two-stage models.

\section{Related Work}
\vspace{-0.5em}
Deep learning-based dense prediction tasks have typically focused on either uncountable background or countable foreground objects.
Semantic segmentation concerns the pixel-wise segmentation of semantics, treating each object class as uncountable.
Instance segmentation, on the other hand, focuses explicitly on countable foreground classes, such as persons or cars.
For the past few years, these tasks have evolved separately, with little interaction,
leading to issues such as trouble with contextual clues in instance segmentation, or
the confusion caused by the large variance of person classes in semantic segmentation.
Recently, panoptic segmentation was proposed as a new task to bridge the gap
between these methods and allow tackling them in a unified manner~\cite{kirillov2019panoptic}.
Embedding-based methods have recently become popular in the computer vision community for improving
object detection~\cite{law2018cornernet,zhou2019objects} and keypoint estimation~\cite{newell2017associative}.
In this section, we briefly review representative methods for each task.

\vspace{-1em}
\paragraph{Semantic segmentation.}
Following the seminal work of Long et al.~\cite{long2015fully}, semantic segmentation is typically treated as a pixel-wise classification task (although exceptions exist~\cite{hwang2019segsort}),
where a fully convolutional network with an encoder-decoder architecture is trained to output a high softmax score for the ground-truth class.
There have been several improvements since then.
Namely, SegNet~\cite{badrinarayanan2017segnet} introduces unpooling layers for more accurate upsampling in the decoder.
Conversely, Deeplab~\cite{chen2017deeplab} proposed to use a network with dilated convolution instead of a decoder, and leverages pooling to capture global information.
PSPNet~\cite{zhao2017pyramid} improves global context by leveraging pyramid pooling and dilated convolution.
DeeplabV3+~\cite{chen2018encoder} combines the advantages of pooling and encoder-decoder architectures to better capture contextual information and sharper object boundaries.

\vspace{-1em}
\paragraph{Instance segmentation.}
Typical high-performing instance segmentation methods are variants of the
Mask R-CNN~\cite{he2017mask} framework~\cite{liu2018path,chen2018masklab}. These methods work in two stages,
where the first stage computes regions of interest via a region proposal network, conducts non-maximum suppression on the proposed bounding boxes, and then runs a second stage via a head network on each proposal.
While yielding high accuracy, these methods are typically too slow for real-time inference.
Recently, there has been work addressing the creation of accurate single-shot (proposal-free) instance segmentation methods~\cite{arnab2017pixelwise,neven2019instance,gao2019ssap,wang2020solo}.
In particular, Neven \etal~\cite{neven2019instance} showed how to accurately predict a spatial embedding space for instance segmentation. Unlike previous methods, their network operates in a single stage and is able to produce
an instance segmentation in a single shot, while having accuracy comparable to the more expensive Mask R-CNN.
Their work inspires us to explore the possibility to learn a hierarchical embedding space for panoptic segmentation.

\vspace{-1.5em}
\paragraph{Panoptic segmentation.}
The current dominant methods for panoptic segmentation are two-stage frameworks.
Kirillov et al.~\cite{kirillov2019panoptic} run PSPNet and Mask R-CNN independently
to obtain semantics and instance predictions. They subsequently combine these using heuristics.
Subsequent work was done to combine the independent networks into one, by adding
a semantic segmentation branch to Mask R-CNN~\cite{kirillov2019panopticfpn,porzi2019seamless,degeus2019panoptic,li2019attention,lazarow2019learning,li2018learning}, but manual heuristics still remained.
Other works aim to further remove manual merging heuristics of the semantics and instance predictions.
E.g. UPSNet~\cite{xiong2019upsnet} proposes a panoptic head network for merging the predictions, and
Liu et al.~\cite{liu2019end} leverages a spatial ranking module to conduct the merging between the two branches.
Yang et al.~\cite{yang2020sognet} propose to resolve overlaps via instance relation reasoning.
Recently, there has been some work adding a mask head on-top of a transformer to avoid heuristics~\cite{carion2020end}.

Single-shot approaches, while less explored~\cite{yang2019deeperlab,gao2019ssap,cheng2020panoptic,hou2020real}, are an important complementary direction for panoptic segmentation. These methods demonstrate their potential in both network accuracy~\cite{cheng2020panoptic} and inference efficiency~\cite{hou2020real}.
These existing methods feature separated feature streams for semantic segmentation- and instance-related representation, while in our approach, a unified embedding is used to model both semantic (category) and instance information. 
That is, our method has in fact \emph{the same} downstream features showing that
this pixel is both \emph{a} car (instance-agnostic), and also \emph{this} car
(instance-specific), creating a natural and inherent feature representation for panoptic segmentation.

\vspace{-1em}
\paragraph{Embeddings for computer vision.}
Associative embeddings~\cite{newell2017pixels,newell2017associative} and variants have been popular for various vision tasks. The embedding space is learned by leveraging push and pull forces between embeddings in the image, depending on ground-truth annotation.
The initial paper on associative embeddings~\cite{newell2017associative} used 1-dimensional embeddings. The follow-up work~\cite{newell2017pixels} showed that increasing the embedding to be 8-dimensional improved convergence of the network.
Embeddings have also been used for lane detection~\cite{neven2018towards}, instance segmentation~\cite{de2017semantic,neven2019instance}, and for instance and semantic segmentation of point clouds~\cite{wang2019associatively}.
Further, embeddings have been used with success for improving object detection methods, using embedding space
as a way to remove human-designed priors~\cite{law2018cornernet,zhou2019objects}.
Our proposed method is a variation of a deep nearest class mean (NCM) classifier~\cite{guerriero2018deep},
which has shown some promising results on image classification for replacing typical networks with softmax outputs directly after the last convolutional layer.
To the best of our knowledge, our work is the first to leverage a deep NCM method for semantics in panoptic segmentation.

\vspace{-0.5em}
\section{Panoptic Segmentation with Hierarchical Embeddings}
\vspace{-0.5em}
In this section, we provide detailed discussion on how we learn the proposed embedding space and use the predicted embeddings for panoptic segmentation, based on the framework depicted in Figure~\ref{fig:method-overview}. We start with problem formulation, and a discussion over the limitation of a current widely used loss in embedding learning for scene parsing. Then we propose the usage of a  better alternative to the task with our novel loss for hierarchical embedding space learning. Finally, we describe panoptic decoding on the learned embedding space for creating a panoptic segmentation.

\subsection{Problem Formulation}
\begin{figure}[t]
{\centering
   \includegraphics[width=1.0\linewidth]{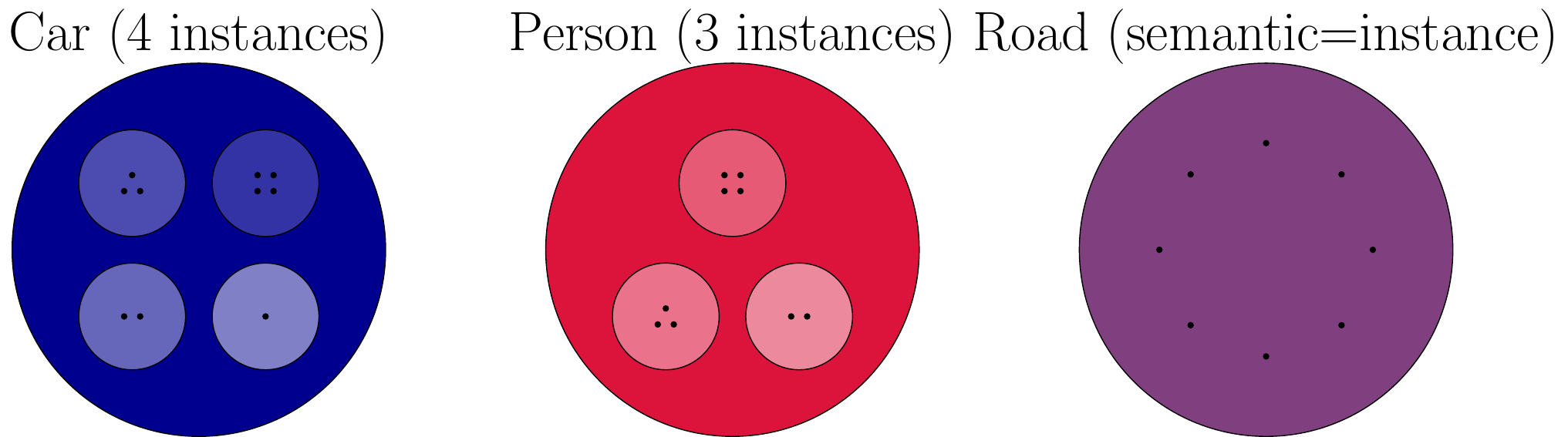}
   \caption{Our proposed loss groups pixel embeddings (black dots) into a hierarchy of jointly semantic- and instance-specific clusters.
   The border of each inner circle illustrates the decision boundary for the instance kernel $\phi$. The outer circle shows the decision boundary for the semantic kernel $\psi$.
   }%
   \label{fig:method-hierarchy}%
}%
\end{figure}

We formulate the task of panoptic segmentation as an embedding space learning problem,
where we want to associate each pixel $\veci{x}{i}$ with a latent embedding $\veci{e}{i}$ such that we can decode
both the object instance and semantic class correctly from \emph{the same} single embedding (cf. Figure~\ref{fig:method-hierarchy}).
Our goal is to learn a function $f$ that maps a set of pixels $\set{X}, \; |\set{X}| = N$, into a set of embeddings $\set{E}$ such that the embeddings can be partitioned into sets $\set{S}$ and $\set{I}$, where $\set{S} = \{S_1, \ldots, S_{|\set{C}|}\}$ are the embedding spaces defined by the set of semantic classes $\set{C}$,
and $\set{I} = \{I_1, \ldots\}$ is the embedding subspace defined by the sets of instances such that we have the hierarchy $I_l \subseteq S_k \subseteq \set{E}$ for any instance $l$ with semantic class $k$. Further, the semantic classes $\set{C}$ are assumed to be divided into thing classes $\set{C}_\text{thing}$ and stuff classes $\set{C}_\text{stuff}$. Note that we consider stuff classes to consist of a single instance.

\subsection{Baseline -- Associative Embeddings}
A popular approach for learning such an embedding is the associative embedding (AE) loss proposed by Newell et al~\cite{newell2017associative} (also known as discriminative loss~\cite{de2017semantic}). The loss uses pull and push terms to
attract or repel an embedding $\veci{e}{i}$ from or to others $\veci{e}{j}$ depending on the ground-truth pixel label of each embedding, given $L$ instances:
\begin{align}
    \mathcal{L}_\text{pull} =& \frac{1}{L} \sum_{l=1}^{L} \frac{1}{|I_l|} \sum_{j \in \set{I}_l} [{\| \veci{\hat{e}}{l} - \veci{e}{j} \|}_1 - \delta_{\text{pull}}]_+^2~, \\
    \mathcal{L}_\text{push} =& \frac{1}{L(L-1)} \sum_{l \neq k} [\delta_{\text{push}} - {\| \veci{\hat{e}}{l} - \veci{\hat{e}}{k} \|}_1]_+^2~,
\end{align}%
\noindent where $\veci{\hat{e}}{l} = \frac{1}{|I_l|} \sum_{j \in \set{I}_l} \veci{e}{j}$ is the instance mean embedding, and $\delta_{\text{pull}}$, $\delta_{\text{push}}$ are problem-specific hinge hyperparameters and $[\cdot]_+$ is the ReLU function.

In theory, a set of two AE losses can be used to learn a hierarchical embedding space by enforcing the hinge hyperparameters to be larger for semantic ground-truth than for instance ground-truth.
While this idea works for toy problems, the disadvantage is that we need to manually define the hinge hyperparameters which employ additional uncertainty and sensitivity in real-world datasets. We have found this difficult to do in practice for real-world datasets such as Cityscapes~\cite{cordts2016cityscapes}, as depicted in our ablation analysis in Section~\ref{sec:ablation}.

\subsection{\Lovasz Hinge Loss}
To decrease the requirement for additional hinge hyperparameters, we look into the \Lovasz hinge loss~\cite{berman2018lovasz}, which acts as a differentiable surrogate for the intersection over union (IoU), a common measure of mask overlap.
In panoptic quality (PQ)~\cite{kirillov2019panoptic} evaluation, IoU is a common base metric across thing and stuff classes.

First, we briefly review the \Lovasz hinge loss.
Given a vector of binary ground-truth labels $\vec{t} \in \{0, 1\}^N$ and predicted labels $\vec{y} \in \{0, 1\}^N$, the IoU is defined as
$\IoU(\vec{y}, \vec{t}) = (|\{i : y_i = 1\} \cap \{i : t_i = 1\}|) / (|\{i : y_i = 1\} \cup \{i : t_i = 1\}|)$,
which is a score between 0 and 1, where higher is better.

As the IoU is a discrete set function, it cannot be optimized via gradient descent, however
the \Lovasz hinge provides a way of creating a continuous and differentiable surrogate function
in terms of the prediction errors $\xi_i$ for each pixel~\cite{berman2018lovasz}.
Given a continuous prediction vector $\vec{p} \in [0,1]^N$, the binary version of the \Lovasz hinge loss is defined as
\begin{align}
    \LovaszBinary(\vec{p}, \vec{t}) = \sum_{i}^N \xi_{\pi_i} \Delta_{i}^\text{IoU}~,
\end{align}%
\noindent with prediction error $\xi$ and IoU difference $\Delta^\text{IoU}$ defined as
\begin{align}
    \xi_i = \begin{cases}
    1 - p_i & t_i = 1 \\
    p_i & t_i = 0
    \end{cases}~, \\
    \Delta_{i}^\text{IoU} = \IoU(\pi_1, \ldots, \pi_{i-1}) - \IoU(\pi_1, \ldots, \pi_i)~, \\
    \pi = \text{indices that sort $\vec{\xi}$ in descending order.} \nonumber
\end{align}%
\noindent Here, with slight abuse of notation, $\IoU(\pi_1, \ldots, \pi_i)$ denotes calculating IoU given the $i$ pixels with the largest prediction error $\xi_{\pi_i}$.
For multi-class prediction, we apply the binary loss in an all-vs-one manner:
\begin{align}
 \resizebox{.90\hsize}{!}{
 {$\LovaszSoftmax(\vec{p}, \vec{t}) = \frac{1}{|C|} \sum_{c \in C} \LovaszBinary(\vec{p}^{c}, \vec{t}^{c})~$,}}
\end{align}%
\noindent where $\vec{t}^{c}$ is 1 if the ground-truth class is $c$, and 0 otherwise.

\begin{figure}[t]
{\centering
   \includegraphics[width=1.0\linewidth]{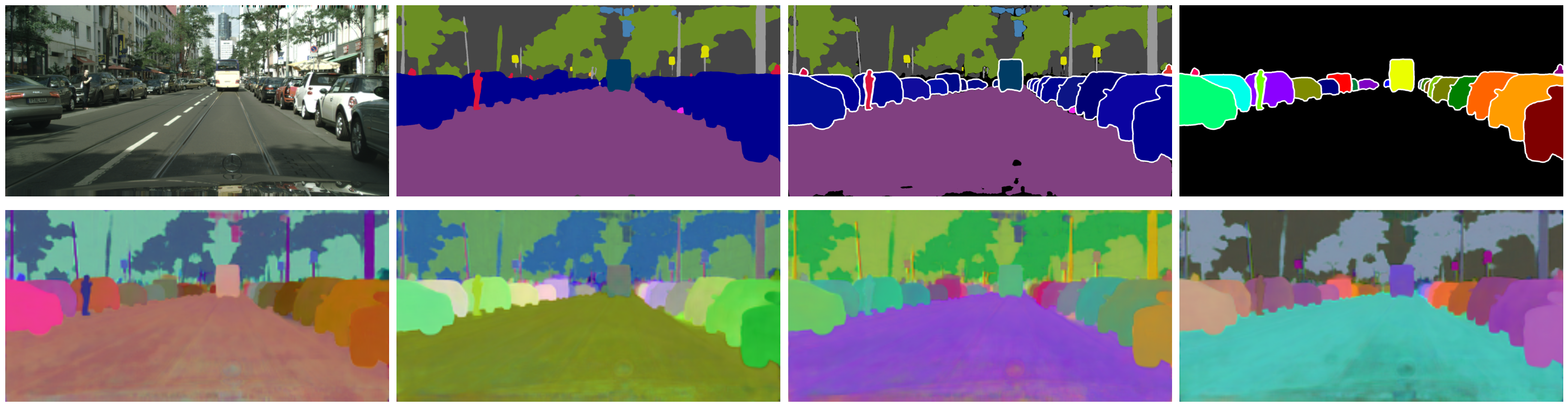}
   \caption{Our method predicts hierarchical embeddings that allow for a clear separation of semantics and instances. Top row from left: input image, semantic segmentation, panoptic segmentation, instance segmentation. Bottom row: our 12-dimensional embedding space visualized as RGB channels of 4 images, highlighting the hierarchical structure over categories and instances.}%
   \label{fig:method-intermediate-outputs}%
}%
\end{figure}%

\subsection{Hierarchical \Lovasz Embeddings}
We propose to leverage the \Lovasz hinge loss to learn a shared embedding space for both semantics and instances.
Given an instance $I_l$, with mean $\veci{\mu}{l}$ and variance $\sigma_l$, the score of an embedding $\veci{e}{i} \in \field{S}^{d-1}$ lying on the unit hypersphere belonging to instance $I_l$ is
\begin{align}
    p_l(\veci{e}{i}) = \exp\left(-\frac{\dcos{\veci{e}{i}}{\veci{\mu}{l}}}{2 \sigma_l^2}\right)~,
\end{align}%
\noindent where $\dcos{\vec{a}}{\vec{b}} = 1 - \vec{a}^\top \vec{b}$ denotes the cosine distance.
Note that $p_l$ plays both roles of the push and pull terms in the AE loss, as embeddings
will be pulled together when the target score increases, and pushed apart otherwise.
We use a unit hypersphere to enable the use of cosine distance, which is memory efficient. However, we emphasize that our method is compatible with any distance metric.

\paragraph{Instance and semantic losses via joint embedding.}
For instances, we use a spatial kernel to better separate far away objects as
\begin{align}
    \phi_l(\veci{e}{i}) = p_l(\veci{e}{i}) \exp\left(-\frac{\| \veci{\rho}{i} - \veci{\rho}{l} \|^2}{2 \sigma_{l,\text{spatial}}^2}\right)~,
\end{align}%
\noindent where $\veci{\rho}{i}$ is the spatial position of embedding $\veci{e}{i}$, and $\sigma_{l,\text{spatial}}$ is a parameter learned by back propagation.

For handling the learning of semantics, we propose to associate with each semantic class $S_k$, a semantic mean $\veci{\mu}{k}$ and semantic variance $\sigma_k$. We can then write the score of an embedding $\veci{e}{i}$ belonging to this semantic class via the softmax function
\begin{align}
    \psi_k(\veci{e}{i}) = \frac{p_k(\veci{e}{i})}{\sum_{c \in \set{C}} p_c(\veci{e}{i})}~.
\end{align}%
\noindent While it is possible to simply use the Gaussian kernel $p_k$ for semantics, preliminary experiments indicated that this is not optimal, as unlike instances, the number of semantic classes are known, and thus the decoding can be done by finding the closest embedding to each semantic mean, which motivates the softmax function.

We define the mean embedding $\veci{\mu}{k}$ to be the mean embedding inside each image according to the ground-truth for instance indices $k$ that exist in the current image, and the persistent estimate $\veci{\hat{\mu}}{k}$ of the dataset mean otherwise:
\begin{align}
    \veci{\mu}{k} = \begin{cases}
    \frac{1}{|I_k|} \sum_{j \in I_k} \veci{e}{j} & \text{if $k$ is an instance} \\
    \veci{\hat{\mu}}{k} & \text{if $k$ is a semantic class}
    \end{cases}~.
\end{align}%

Given the predicted instance and semantic scores $\phi_l(\veci{e}{i})$ and $\psi_k(\veci{e}{i})$ for each embedding $\veci{e}{i}$, we can minimize the \Lovasz hinge loss, to maximize the IoU metric on the training dataset. Our loss functions that act as the push and pull forces to form the hierarchical structure are defined as
\begin{align}
    \mathcal{L}_\text{ins} = \LovaszBinary(\phi_l(\veci{e}{i}), \vec{t}_\text{ins}),~ \\
    \mathcal{L}_\text{seg} = \LovaszSoftmax(\psi_k(\veci{e}{i}), \vec{t}_\text{seg})~,
\end{align}%
\noindent which are calculated for all embeddings $\veci{e}{i}$, instances $I_l$ and semantics $S_k$ and the ground-truth vectors $\vec{t}$.
This will cause the network to pull embeddings belonging to the same class together, and push apart embeddings that belong to different instances. The instance-semantic hierarchy is modeled by letting the network learn variance estimates that properly capture the hierarchy in the shared embedding space, with the semantic variance being larger than the instance variance.

\paragraph{Auxiliary losses.}
In addition to the embeddings for each pixel, our network also predicts a sigmoid output $s_i$ for each pixel, estimating the value of the Gaussian kernel $\phi_l(\veci{e}{i})$.
As shown in previous work, this ``seed'' value can be used at the decoding
step to help finding an arbitrary number of instances~\cite{neven2019instance}. For each pixel, we calculate
\begin{align}
    \mathcal{L}_\text{seed} = 
    \begin{cases}
    \| s_i - \sg{\phi_l(\veci{e}{i})} \|^2 & i \in I_l \\
    \| s_i \|^2 & c_i \in \set{C}_\text{stuff} \\
    \end{cases}~,
\end{align}%
\noindent where $\sg{\cdot}$ is the stop gradient operator, e.g. as used in VQ-VAE~\cite{razavi2019generating}.
Note that only instances are predicted to have a seed value, and stuff classes are regressed to zero.

We also add a loss term for each embedding $\veci{e}{i} \in \set{I}$ to encourage the predicted instance variances to be similar for the same object, to ease instance separation:
\begin{align}
    \mathcal{L}_\text{ins-var} = \gamma\| \sigma_i - \sg{\hat{\sigma}_l} \|^2~,
\end{align}%
\noindent where $\hat{\sigma}_l = \frac{1}{|I_l|} \sum_{j \in I_l} \sigma_j$,
to encourage uniform predicted variance, and $\gamma=10$.

Similar to previous work by Guerriero et al.~\cite{guerriero2018deep}, we found that the semantic means are difficult to learn via backpropagation, as the update is too slow to catch up with the other parts of the network.
Therefore, similar to VQ-VAE~\cite{razavi2019generating}, for semantic classes, we add an L2 regression loss term to store a persistent representation
of each semantic class in the model
\begin{align}
    \mathcal{L}_\text{seg-mean} = \| \veci{\hat{\mu}}{k} - \sg{\veci{\mu}{k}^\text{batch}} \|^2~,
\end{align}%
\noindent where $\veci{\mu}{k}^\text{batch}$ is the batch semantic mean gotten by averaging the predicted embeddings at the ground-truth locations of semantic class $k$.
We illustrate the effect of this VQ-VAE style loss term in our ablation analysis in Section~\ref{sec:ablation}.
The semantic variance $\sigma_k$ is learned via backpropagation.

\paragraph{Proposed loss.}
Our proposed loss function is independent of the specific network architecture, and can be used with any standard fully convolutional neural network for semantic segmentation, such as DeeplabV3+~\cite{chen2018encoder} or PSPNet~\cite{zhao2017pyramid}.
The final form of our proposed loss function is
\begin{align}
    \mathcal{L} = \mathcal{L}_\text{seg} + \mathcal{L}_\text{seg-mean} + \mathcal{L}_\text{ins} + \mathcal{L}_\text{ins-var} + \mathcal{L}_\text{seed}~,
\end{align}%
\noindent averaged over all pixels $\set{X}$.
Specifically, our network predicts, for each pixel $\veci{x}{i}$, a tuple $(\veci{e}{i}, \sigma_i, s_i)$, where $\veci{e}{i}$ is the predicted embedding, $\sigma_i$ is the predicted instance variance, and $s_i$ is the seed probability score.

\paragraph{Thomson initialization.}
To increase coverage of the embedding space, we can initialize the means to uniformly cover the unit hypersphere by solving the generalized Thomson problem~\cite{thomson1904structure}, originally proposed in 1904 for the purpose of determining an electron configuration in physics.
We initialize the means as
\begin{align}
    \argmin_{\{\veci{\mu}{k}\}_{\forall k}} \sum_{i \neq j} \frac{1}{d_\text{cos}(\veci{\mu}{i}, \veci{\mu}{j})}~.
\end{align}%
There is no known general solution to this problem, but we can find a local minimum via gradient descent. In our initial experiments, we found that Thomson initialization makes the training more stable.

\subsection{Panoptic Decoding}
\label{sec:decoding}
For predicting semantics and instances from the learned hierarchical embedding space,
we use a simple post-processing algorithm.
We first assign the semantic class to each pixel via argmax of $\psi_k(\veci{e}{i})$.
For thing classes, we first reduce the number of seed candidates $s_i$ by $3 \times 3$ max pooling, similar to CornerNet~\cite{law2018cornernet}.
Then we threshold the remaining seeds to get an initial set of candidate seeds.
Each candidate seed $s_i$ has an associated hierarchical embedding $\veci{e}{i}$,
so we can use the  kernel $\Phi(\veci{e}{i}, \veci{e}{j}) = \exp(-\dcos{\veci{e}{i}}{\veci{e}{j}}/ (2 \sigma_i^2) -\| \veci{\rho}{i} - \veci{\rho}{j} \|^2/(2 \sigma_{i,\text{spatial}}^2))$ to calculate the probability of two seeds $(s_i, s_j)$ representing the same instance.
We then merge seeds representing the same object based on the probability $\Phi(\veci{e}{i}, \veci{e}{j})$. We now have one estimated seed $s_i$ per instance.
For each $s_i$, we then estimate the instance mask by thresholding $\Phi(\veci{e}{i}, \veci{e}{k})$, for all remaining pixel embeddings $\veci{e}{k}$, resolving mask disagreement by assigning each pixel to the instance with the highest probability among the estimated seeds. The semantic class of the instance is decided by the seed pixel.
For stuff classes, we threshold the value of the semantic kernel $\psi_k(\veci{e}{i})$, to reduce the number of false positives.
The panoptic decoding process can be easily optimized in favor of inference speed with little degradation of accuracy (details in the supplementary).

\setlength{\tabcolsep}{1.4pt}
\begin{table}[t]
\centering
\begin{tabular}{|l|l|l|c|c|c|}
\hline
Method & Backbone & Pretrain. & $PQ$ & $PQ_{th}$ & $PQ_{st}$ \\
\hline
\multicolumn{6}{l}{Proposal-based}  \\
\hline
Seamless~\cite{porzi2019seamless} & ResNet50 & ImageNet & $\mathbf{60.3}$ & $\mathbf{56.1}$ & $63.3$ \\
UPSNet~\cite{xiong2019upsnet} & ResNet50 & ImageNet & 59.3 & 54.6 & 62.7 \\  %
Real-time PS~\cite{hou2020real} & ResNet50 & ImageNet & 58.8 & 52.1 & $\mathbf{63.7}$ \\
Pan.FPN~\cite{kirillov2019panopticfpn} & ResNet50 & ImageNet & 57.7 & 51.6 & 62.2 \\
Attn.-Guid.~\cite{li2019attention} & ResNet50 & ImageNet & 56.4 & 52.7 & 59.0 \\
Li et al.~\cite{li2018weakly} & ResNet101 & ImageNet & 47.3 & 39.6 & 52.9 \\
DeGeus~\cite{degeus2019panoptic} & ResNet50 & ImageNet & 45.9 & 39.2 & 50.8 \\
\hline
\multicolumn{6}{l}{Proposal-free} \\
\hline
Pan. DeepL.~\cite{cheng2020panoptic} & ResNet50 & ImageNet & $59.7$ & - & - \\
SSAP~\cite{gao2019ssap} & ResNet50 & ImageNet & 57.6 & $50.4$ & - \\
DeeperLab~\cite{yang2019deeperlab} & Xception71 & ImageNet & 56.5 & - & - \\
DeGeusFast~\cite{degeus2019fast} & ResNet50 & ImageNet & 55.1 & 48.3 & 60.1 \\
\hline
\textbf{HLE (Ours)}& ResNet101 & ImageNet & $60.6$ & $51.4$ & $67.2$ \\
\textbf{HLE (Ours)}& ResNet50 & ImageNet & $\mathbf{59.8}$ & $\mathbf{51.1}$ & $\mathbf{66.1}$ \\
\textbf{HLE (Ours)}& MobileNetV2 & ImageNet & $56.0$ & $45.0$ & $64.0$ \\
\hline
\end{tabular}
\vspace{0.5em}
\caption{Single-scale experimental results on the Cityscapes validation set. The best ResNet50 result in each category is highlighted.}
\label{tbl:cityscapes-val}
\vspace{-1.0em}
\end{table}%
\setlength{\tabcolsep}{1.4pt}

\section{Experiments}
\subsection{Datasets}
We conduct experiments on the Cityscapes~\cite{cordts2016cityscapes}, COCO~\cite{lin2014microsoft}, and Mapillary Vistas~\cite{neuhold2017mapillary} datasets, to evaluate the performance of our model.
Cityscapes is an image dataset for autonomous driving, depicting European street-level imagery at $1024 \times 2048$ resolution, labeled with 19 classes, out of which 8 are thing classes.
COCO contains various indoor and outdoor images at varying resolutions, with 80 thing classes, and 53 stuff classes.
Vistas contains a wide variety of street-level images, at varying resolution, with 37 thing classes and 28 stuff classes.

\subsection{Experimental Details}
We evaluate our method using the standard panoptic segmentation metric panoptic quality (PQ)~\cite{kirillov2019panoptic}. In the supplementary material, we explain the $PQ$ metric, and include its variations, $PQ^\dagger$~\cite{porzi2019seamless} and parsing covering (PC)~\cite{yang2019deeperlab}, as well as mean Intersection-over-Union (mIoU) and Average Precision (AP).

On Cityscapes, we only use the fine annotations in training. Our models are trained
on the training set ($2,975$ images) and evaluated on the validation set ($500$ images).
For COCO, the training and validation sets have $118,287$ and $5,000$ images, respectively.
Vistas has $18,000$ images in the training set, and $2,000$ in the validation set.
We use the Adam optimizer~\cite{kingma2014adam} with learning rate $10^{-5}$, polynomial learning rate decay, and test-time flip.
We use ResNet50 as a backbone for DeepLabV3+ (pretrained on ImageNet) with an embedding dimension of 12 on Cityscapes, and 128 on COCO and Vistas.
During training, we use crop size $1024 \times 2048$ on Cityscapes, and $512 \times 512$ on COCO, and crop around thing classes.
The experiments on COCO and Vistas uses a max side length of 640 and 2048 pixels, respectively.
We use Optuna~\cite{akiba2019optuna} to find good hyperparameters for the decoder.

Network inference on a $1024 \times 2048$ image takes $91$ ms running on single NVIDIA V100 Tensor Core GPU. Post-processing takes $113$ ms. Alternatively, by running post-processing on a down-sampled embedding space, post-processing speed can be easily sped up to $8$ ms at the expense of lowering Cityscapes validation set PQ to $58.4$ (details are provided in the supplementary material). On a $800 \times 1300$ COCO image, inference takes $51$ ms. Inference speed depends heavily on the backbone used. With a MobileNetV2 backbone, model inference takes 36 ms on Cityscapes.

\setlength{\tabcolsep}{4pt}
\begin{table}[t]
\centering
\begin{tabular}{|l|l|l|c|}
\hline
Method & Backbone & Pretrain. & $PQ$ \\
\hline
\multicolumn{4}{l}{Proposal-based}  \\
\hline
Seamless~\cite{porzi2019seamless} & ResNet50 & ImageNet \& Vistas & $\mathbf{62.6}$ \\
\hline
\multicolumn{4}{l}{Proposal-free} \\
\hline
SSAP~\cite{gao2019ssap} & ResNet101 & ImageNet & $58.9$ \\
Dynam. inst.~\cite{arnab2017pixelwise} & ResNet101 & ImageNet & 55.4 \\
\hline
\textbf{HLE (Ours)}& ResNet101 & ImageNet & $\mathbf{59.4}$ \\
\textbf{HLE (Ours)}& ResNet50 & ImageNet & $57.9$ \\
\hline
\end{tabular}
\vspace{0.5em}
\caption{Experimental results on the Cityscapes test set.
}
\label{tbl:cityscapes-test}
\vspace{-0.75em}
\end{table}%
\setlength{\tabcolsep}{1.4pt}

\setlength{\tabcolsep}{1.4pt}
\begin{table}[t]
\centering
\begin{tabular}{|l|l|l|c|c|c|}
\hline
Method & Backbone & Pretrain. & $PQ$ & $PQ_{th}$ & $PQ_{st}$ \\
\hline
\multicolumn{6}{l}{Proposal-based}  \\
\hline
UPSNet~\cite{xiong2019upsnet} & ResNet50 & ImageNet & $\mathbf{42.5}$ & $48.5$ & $\mathbf{33.4}$ \\  %
Attn.-Guid.~\cite{li2019attention} & ResNet50 & ImageNet & 39.6 & $\mathbf{49.1}$ & 25.2 \\
Pan. FPN~\cite{kirillov2019panopticfpn} & ResNet50 & ImageNet & 39.0 & 45.9 & 28.7 \\
Real-time PS~\cite{hou2020real} & ResNet50 & ImageNet & 37.1 & 41.0 & 31.3 \\
\hline
\multicolumn{6}{l}{Proposal-free} \\
\hline
SSAP~\cite{gao2019ssap} & ResNet101 & ImageNet & $36.5$ & - & - \\
Pan. DeepL.~\cite{cheng2020panoptic} & ResNet50 & ImageNet & $35.1$ & - & - \\
DeeperLab~\cite{yang2019deeperlab} & Xception71 & ImageNet & 33.8 & - & - \\
\hline
\textbf{HLE (Ours)}& ResNet101 & ImageNet & $38.1$ & $42.8$ & $31.0$ \\
\textbf{HLE (Ours)}& ResNet50 & ImageNet & $\mathbf{37.1}$ & $\mathbf{41.1}$ & $\mathbf{30.9}$ \\
\hline
\end{tabular}
\vspace{0.5em}
\caption{Single-scale experimental results on the COCO validation set. The best ResNet50 result in each category is highlighted.}
\label{tbl:coco-val}
\vspace{-1.5em}
\end{table}%
\setlength{\tabcolsep}{1.4pt}
\setlength{\tabcolsep}{1.4pt}
\begin{table}[ht]
\centering
\begin{tabular}{|l|l|l|c|c|c|}
\hline
Method & Backbone & Pretrain. & $PQ$ & $PQ_{th}$ & $PQ_{st}$ \\
\hline
\multicolumn{6}{l}{Proposal-based}  \\
\hline
UPSNet~\cite{xiong2019upsnet} & ResNet101 & ImageNet & $\mathbf{46.6}$ & 53.2 & $\mathbf{36.7}$ \\
Attn.-Guid.~\cite{li2019attention} & ResNeXt-152 & ImageNet & 46.5 & $\mathbf{55.8}$ & 32.5 \\
Pan. FPN~\cite{kirillov2019panopticfpn} & ResNet50 & ImageNet & 40.9 & 48.3 & 29.7 \\
\hline
\multicolumn{6}{l}{Proposal-free} \\
\hline
SSAP~\cite{gao2019ssap} & ResNet101 & ImageNet & $36.9$ & $40.1$ & $32.0$ \\
DeeperLab~\cite{yang2019deeperlab} & Xception71 & ImageNet & 34.3 & 37.5 & 29.6 \\
DeeperLab~\cite{yang2019deeperlab} & Wider MNV2 & ImageNet & 28.1 & 30.8 & 24.1 \\
DeeperLab~\cite{yang2019deeperlab} & L. W. MNV2 & ImageNet & 24.5 & 26.9 & 20.9 \\
\hline
\textbf{HLE (Ours)}& ResNet101 & ImageNet & $\mathbf{39.9}$ & $\mathbf{45.0}$ & $\mathbf{32.2}$ \\
\textbf{HLE (Ours)}& ResNet50 & ImageNet & $38.2$ & $42.7$ & $31.4$ \\
\hline
\end{tabular}
\vspace{0.5em}
\caption{Experimental results on the COCO test-dev2019 set.
}
\label{tbl:coco-testdev2017}
\end{table}%
\setlength{\tabcolsep}{1.4pt}
\begin{table}[t]
\centering
\begin{tabular}{|l|l|l|c|c|c|}
\hline
Method & Backbone & Pretrain. & $PQ$ & $PQ_{th}$ & $PQ_{st}$ \\
\hline
\multicolumn{6}{l}{Proposal-based}  \\
\hline
Seamless~\cite{porzi2019seamless} & ResNet50 & ImageNet & $\mathbf{37.7}$ & $\mathbf{33.8}$ & $\mathbf{42.9}$ \\
\hline
\multicolumn{6}{l}{Proposal-free} \\
\hline
Pan. DeepL.~\cite{cheng2020panoptic} & ResNet50 & ImageNet & $33.3$ & - & - \\
DeeperLab~\cite{yang2019deeperlab} & Xception71 & ImageNet & $32.0$ & - & - \\
\hline
\textbf{HLE (Ours)}& ResNet50 & ImageNet & $\mathbf{34.6}$ & $\mathbf{27.8}$ & $\mathbf{43.5}$ \\
\hline
\end{tabular}
\vspace{0.5em}
\caption{Single-scale experimental results on the Vistas validation set. The best ResNet50 result in each category is highlighted.}
\label{tbl:vistas-val}
\end{table}%
\setlength{\tabcolsep}{1.4pt}

\begin{figure}[t]
{\centering
\def\figwidth{0.45}
\includegraphics[width=\figwidth\linewidth]{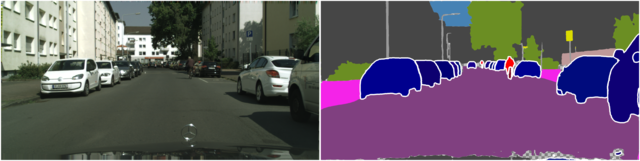}
\includegraphics[width=\figwidth\linewidth]{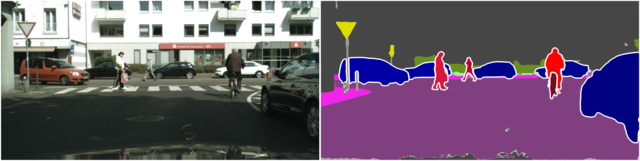}\\
\includegraphics[width=\figwidth\linewidth]{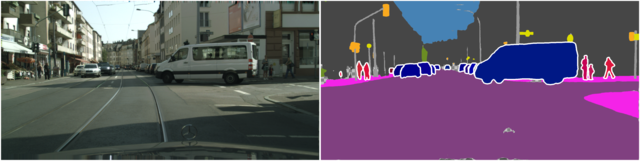}
\includegraphics[width=\figwidth\linewidth]{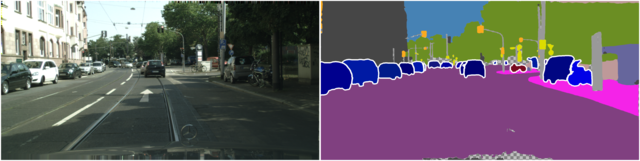}\\
\includegraphics[width=\figwidth\linewidth]{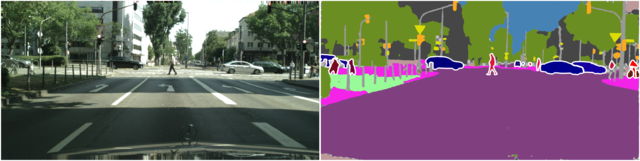}
\includegraphics[width=\figwidth\linewidth]{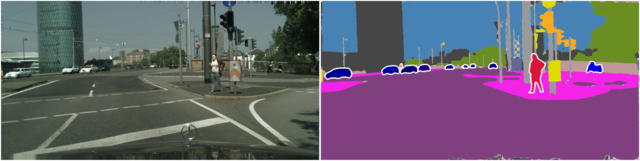}
   \caption{Visualization of our method on the Cityscapes validation set.
   Each pair: (left) input image, (right) panoptic segmentation.
   }%
   \label{fig:cityscapes-val-visualization}%
}%
\end{figure}%

\begin{figure}[t]
{\centering
\def\figwidth{0.2}
\def\figheight{2.6em}
\includegraphics[height=\figheight,keepaspectratio]{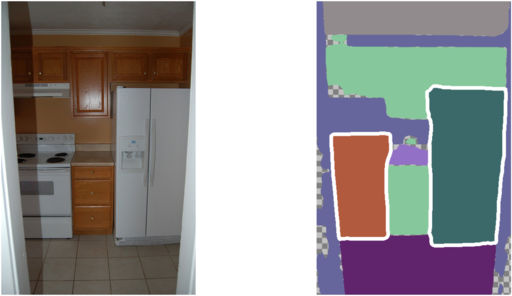}
\includegraphics[height=\figheight,keepaspectratio]{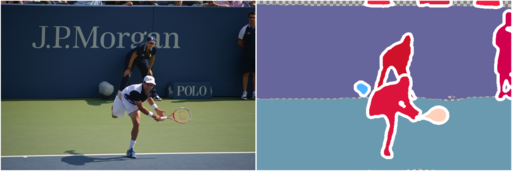}
\includegraphics[height=\figheight,keepaspectratio]{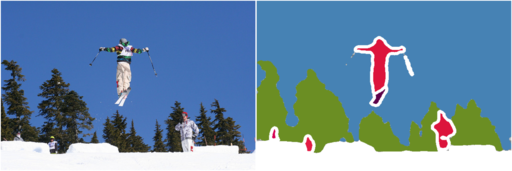}\\
\includegraphics[height=\figheight,keepaspectratio]{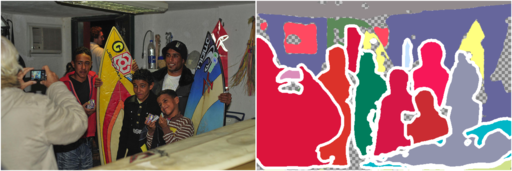}
\includegraphics[height=\figheight,keepaspectratio]{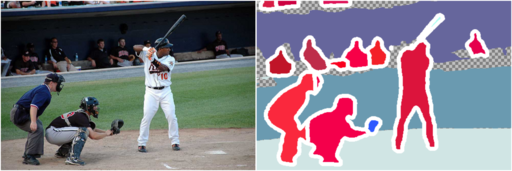}
\includegraphics[height=\figheight,keepaspectratio]{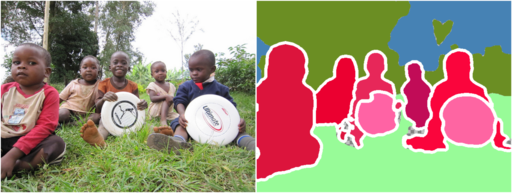}
   \caption{Visualization of our method on the COCO validation set.
   Each pair: (left) input image, (right) panoptic segmentation.
   }%
   \label{fig:coco-val-visualization}%
}%
\vspace{-1.0em}
\end{figure}%

\begin{figure}[t]
{\centering
\def\figwidth{0.45}
\includegraphics[width=\figwidth\linewidth]{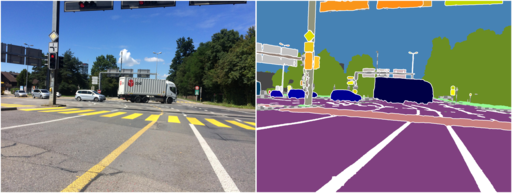}
\includegraphics[width=\figwidth\linewidth]{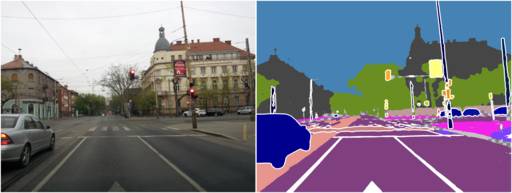}\\
\includegraphics[width=\figwidth\linewidth]{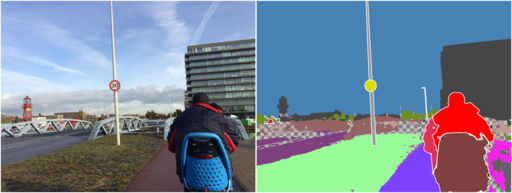}
\includegraphics[width=\figwidth\linewidth]{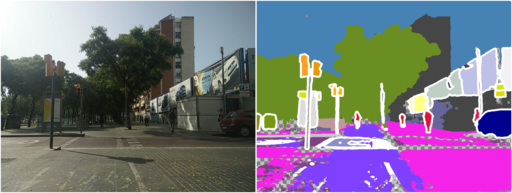}\\
\includegraphics[width=\figwidth\linewidth]{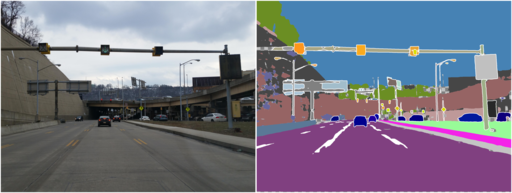}
\includegraphics[width=\figwidth\linewidth]{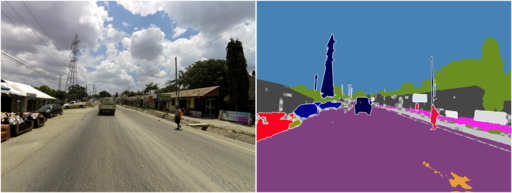}
   \caption{Visualization of our method on the Vistas validation set.
   Each pair: (left) input image, (right) panoptic segmentation.
   }%
   \label{fig:vistas-val-visualization}%
}%
\end{figure}%

\setlength{\tabcolsep}{2pt}
\begin{table}[t]
\centering
\begin{tabular}{|l|c|c|c|c|c|c||c|}
\hline
Assoc. emb. loss & \checkmark & & & & & & \\
Cross-ent. loss & & \checkmark & & \checkmark & & & \\
\Lovasz hinge loss & & & \checkmark & \checkmark & \checkmark & \checkmark & \checkmark \\
\hline
Sep. seg. branch. & & & & \checkmark & & & \\
Split emb. space & & & \checkmark & & & & \\
Hier. emb. space & \checkmark & \checkmark & & & \checkmark & \checkmark & \checkmark \\ \hline
VQ-VAE style & & \checkmark & \checkmark & & & \checkmark & \checkmark \\ \hline
Thomson init. & & \checkmark & \checkmark & \checkmark & \checkmark & & \checkmark \\ \hline
PQ & $38.8$ & $45.0$ & $50.2$ & $57.0$ & $57.0$ & $58.6$ & $\mathbf{59.1}$ \\
\hline
\end{tabular}
\vspace{0.5em}
\caption{Ablative results on the Cityscapes validation set. Our proposed model is in the rightmost column.}
\label{tbl:cityscapes-val-ablative}
\vspace{-1.0em}
\end{table}%
\setlength{\tabcolsep}{1.4pt}

\subsection{Experimental Results}
Our results on Cityscapes, COCO, and Vistas can be seen in Tables~\ref{tbl:cityscapes-val},~\ref{tbl:coco-val} and~\ref{tbl:vistas-val}, respectively.
The best ResNet50 result in each category is highlighted.
Some examples of our model's outputs can be seen in Figures~\ref{fig:cityscapes-val-visualization},~\ref{fig:coco-val-visualization} and~\ref{fig:vistas-val-visualization}.
For a fair comparison, we report the results on ImageNet-trained models with a ResNet50 for previous works, if possible.
We also report some results of our method with alternative backbones~\cite{he2016deep,sandler2018mobilenetv2}, to illustrate the change in performance depending on the backbone network.
Our method achieves state-of-the-art results among proposal-free methods on all three datasets with the same or even lighter backbone.

On Cityscapes, we can see that our method is competitive with heavier two-stage methods, such as PanopticFPN~\cite{kirillov2019panopticfpn}, and has higher $PQ$ than previous proposal-based methods, such as Panoptic DeepLab~\cite{cheng2020panoptic}.
Particularly, note that our method with a ResNet50 backbone network beats DeeperLab that uses a much heavier Xception71 backbone in terms of the $PQ$ metric.

Notably, we achieve a high $PQ_\text{st}$ score, indicating that using embeddings for semantic encoding is a promising direction, especially in panoptic segmentation.
We can reason about the rationale that our model achieves top $PQ_\text{st}$ results as follows. In our method, we learn both embeddings and variances for each semantic class. Each semantic class is treated in an adaptive manner based on distribution. For example, a larger volume in the embedding space will be assigned to semantic classes with higher variance, compared to direct regression where all classes are treated equally in feature space.
An example of the learned hierarchical embedding space can be seen in Figure~\ref{fig:method-intermediate-outputs}.
Our proposed hierarchical embedding is able to encode both semantic and instance level information precisely, although a few mistaken predictions can also be seen.

We also submitted our model to the Cityscapes test set benchmark, where our method is competitive among both published and unpublished models, despite the big variety in network backbones and extra data included in the current benchmark.
The results for peer-reviewed work can be seen in Table~\ref{tbl:cityscapes-test}.
The gap between our validation and test set $PQ$ score is not large, indicating that our method generalizes well. 
While SSAP does not report any ResNet50 result, our model has higher PQ than theirs with ResNet101.
In the supplementary material, we include Cityscapes results comparison on the metrics $PQ^\dagger$ ($61.3$), $PC$ ($76.6$), $mIoU$ ($77.3$) and $AP$ ($23.9$).

On COCO, we saw less proposal-free methods reported with similar network scale as shown in Table~\ref{tbl:coco-val}.
We present COCO test set results in Table~\ref{tbl:coco-testdev2017}.
Our method achieves the best performance among proposal-free models.
Compared to Cityscapes, the gap between proposal-based and proposal-free models on COCO is much larger, indicating that there is still a demand for further research in order to close this gap.
We find the same observation on the Vistas dataset as on COCO, where our proposed method is able to outperform other proposal-free methods.
Notably, we find that bottom-up methods tend to under-perform on larger datasets~\cite{cheng2020panoptic,hou2020real,gao2019ssap}, while our proposed hierarchical 
embedding space-based method is able to generalize better on these datasets.

\vspace{-0.25em}
\subsection{Ablative Analysis}
\vspace{-0.25em}
\label{sec:ablation}
We conduct ablative experiments to better understand design choices we have made for our method.
In Table~\ref{tbl:cityscapes-val-ablative}, we provide different variations on each key design of our proposed method. For simplicity, we did not conduct hyper-parameter search in this experiment.

First of all, we compared our proposed hierarchical \Lovasz embeddings with conventional embeddings based on associative embedding or cross entropy loss.
As can be seen in the table, \Lovasz embeddings have vastly better performance than associative embeddings or cross entropy losses.

Second, we compare with a model that uses a softmax classifier semantic segmentation branch with cross entropy loss, instead of our joint hierarchical embedding space, and still handling instance predictions via the embedding space. We can see that this yields a lower PQ of $57.0$, which can be regarded as a na\"ive extension of previous instance segmentation work to panoptic segmentation~\cite{neven2019instance}.
Next, we illustrate the benefit of jointly learning the embedding space for both semantics and instances, compared to splitting the embedding space into two halves: one for semantics and the other for instance embeddings.
We split the 12-dimensional embedding space of our model into 6 dimensions each for semantics and instances. It can be seen that jointly learning both semantics and instances in the same embedding space leads to higher performance.
This interpretation is also strengthened by our method outperforming DeeperLab on the Cityscapes validation set -- a method using a split embedding space.

Finally, we show that VQ-VAE style loss and Thomson initialization for learning the semantic means help improve the model performance.

\begin{figure}[t]
{\centering
\begin{minipage}{0.5\textwidth}
\def\figwidth{0.8}
$t=0$\hphantom{0} \includegraphics[width=\figwidth\linewidth]{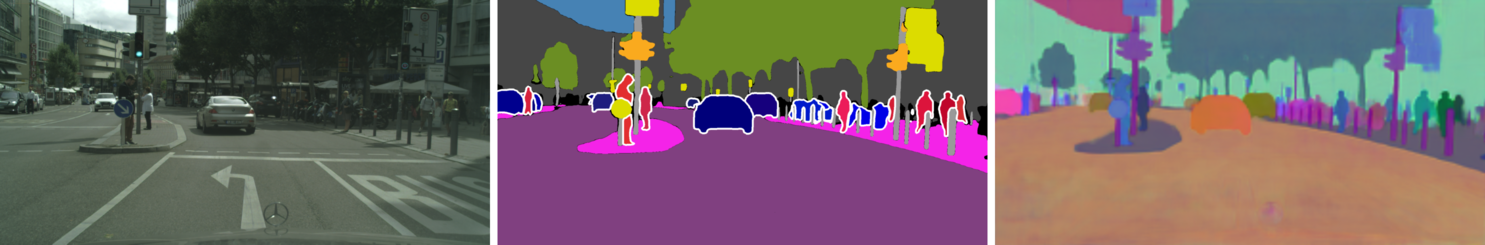}\\
$t=5$\hphantom{0} \includegraphics[width=\figwidth\linewidth]{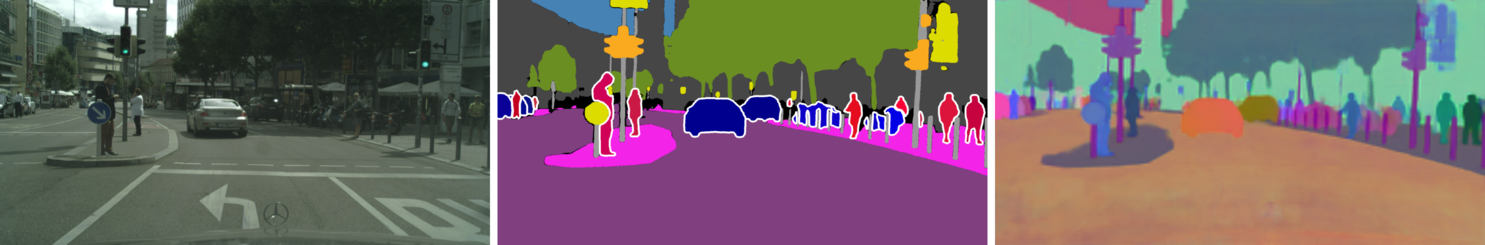}\\
$t=10$ \includegraphics[width=\figwidth\linewidth]{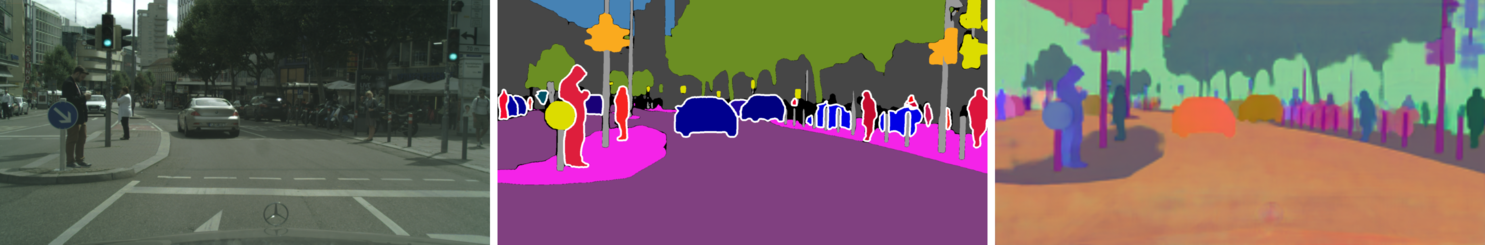}\\
$t=15$ \includegraphics[width=\figwidth\linewidth]{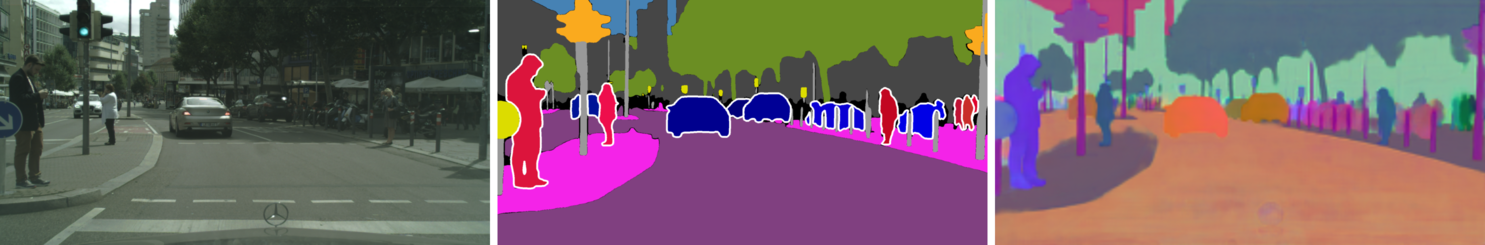}
\vspace{1.0em}
\end{minipage}
   \caption{Our embeddings are temporally stable across video frames. Each row (video frame), from left to right: Input image, panoptic segmentation, first 3 dimensions of the embedding space. Matching pixels in consecutive frames have similar embeddings.}%
   \label{fig:cityscapes-temporal-stability}%
}%
\vspace{-0.5em}
\end{figure}%

\subsection{Temporal Stability}
\label{sec:temporal-stability}
\vspace{-0.5em}
As our embeddings are of low dimension, we are interested in seeing whether they exhibit temporal stability when run on a video. In Figure~\ref{fig:cityscapes-temporal-stability}, we show the predicted embedding spaces and panoptic segmentation for a few consecutive frames in the Cityscapes demo video sequence. The figure qualitatively shows that matching pixels in different frames have similar embeddings. We believe this qualitative result indicates the potential of our method towards temporal downstream applications such as multi-object tracking.
Temporal stability is important for applications such as tracking in autonomous driving, where flickering false positives can be fatal, and low-dimensional embeddings could be leveraged as a feature for cross-frame data association.
We are not aware of other panoptic methods featuring this property.

\section{Conclusion}
In this work, we presented a unified output representation for panoptic segmentation leveraging a hierarchically clustered embedding space.
Via our proposed hierarchical \Lovasz hinge loss, we created a simple single-shot model for panoptic segmentation, predicting an embedding space structured into an instance-semantic hierarchy.
Our method was shown to achieve state-of-the-art results among proposal-free panoptic segmentation methods,
and be competitive with heavier two-stage methods on the Cityscapes, COCO and Vistas datasets.
Results indicated that our hierarchical \Lovasz embeddings are a viable alternative for panoptic segmentation, for both thing and stuff classes.

We believe that it is an important and promising direction to explore a unified representation for panoptic segmentation. In the future, we will focus on leveraging the structure of the embedding space with self-adaptation to ontology distribution and its downstream applications.

{
\vfill
\paragraph{Acknowledgments}
This work was supported by Woven Core, Inc.
We thank Richard Calland, Aditya Ganeshan, Karim Hamzaoui, Daisuke Okanohara and Shunta Saito for discussions and feedback on the manuscript.
}
\clearpage
\balance
{\small
\bibliographystyle{ieee_fullname}
\bibliography{paper}

\begin{thebibliography}{10}\itemsep=-1pt

\bibitem{akiba2019optuna}
Takuya Akiba, Shotaro Sano, Toshihiko Yanase, Takeru Ohta, and Masanori Koyama.
\newblock Optuna: A next-generation hyperparameter optimization framework.
\newblock In {\em Proceedings of the 25th ACM SIGKDD International Conference
  on Knowledge Discovery \& Data Mining}, pages 2623--2631. ACM, 2019.

\bibitem{arnab2017pixelwise}
Anurag Arnab and Philip~HS Torr.
\newblock Pixelwise instance segmentation with a dynamically instantiated
  network.
\newblock In {\em Proceedings of the IEEE Conference on Computer Vision and
  Pattern Recognition}, pages 441--450, 2017.

\bibitem{badrinarayanan2017segnet}
Vijay Badrinarayanan, Alex Kendall, and Roberto Cipolla.
\newblock Segnet: A deep convolutional encoder-decoder architecture for image
  segmentation.
\newblock {\em IEEE transactions on pattern analysis and machine intelligence},
  39(12):2481--2495, 2017.

\bibitem{berman2018lovasz}
Maxim Berman, Amal Rannen~Triki, and Matthew~B Blaschko.
\newblock The lov{\'a}sz-softmax loss: A tractable surrogate for the
  optimization of the intersection-over-union measure in neural networks.
\newblock In {\em Proceedings of the IEEE Conference on Computer Vision and
  Pattern Recognition}, pages 4413--4421, 2018.

\bibitem{carion2020end}
Nicolas Carion, Francisco Massa, Gabriel Synnaeve, Nicolas Usunier, Alexander
  Kirillov, and Sergey Zagoruyko.
\newblock End-to-end object detection with transformers.
\newblock In {\em European Conference on Computer Vision}, pages 213--229.
  Springer, 2020.

\bibitem{chen2018masklab}
Liang-Chieh Chen, Alexander Hermans, George Papandreou, Florian Schroff, Peng
  Wang, and Hartwig Adam.
\newblock Masklab: Instance segmentation by refining object detection with
  semantic and direction features.
\newblock In {\em Proceedings of the IEEE Conference on Computer Vision and
  Pattern Recognition}, pages 4013--4022, 2018.

\bibitem{chen2017deeplab}
Liang-Chieh Chen, George Papandreou, Iasonas Kokkinos, Kevin Murphy, and Alan~L
  Yuille.
\newblock Deeplab: Semantic image segmentation with deep convolutional nets,
  atrous convolution, and fully connected crfs.
\newblock {\em IEEE transactions on pattern analysis and machine intelligence},
  40(4):834--848, 2017.

\bibitem{chen2018encoder}
Liang-Chieh Chen, Yukun Zhu, George Papandreou, Florian Schroff, and Hartwig
  Adam.
\newblock Encoder-decoder with atrous separable convolution for semantic image
  segmentation.
\newblock In {\em Proceedings of the European conference on computer vision
  (ECCV)}, pages 801--818, 2018.

\bibitem{cheng2020panoptic}
Bowen Cheng, Maxwell~D Collins, Yukun Zhu, Ting Liu, Thomas~S Huang, Hartwig
  Adam, and Liang-Chieh Chen.
\newblock Panoptic-deeplab: A simple, strong, and fast baseline for bottom-up
  panoptic segmentation.
\newblock In {\em Proceedings of the IEEE/CVF Conference on Computer Vision and
  Pattern Recognition}, pages 12475--12485, 2020.

\bibitem{cordts2016cityscapes}
Marius Cordts, Mohamed Omran, Sebastian Ramos, Timo Rehfeld, Markus Enzweiler,
  Rodrigo Benenson, Uwe Franke, Stefan Roth, and Bernt Schiele.
\newblock The cityscapes dataset for semantic urban scene understanding.
\newblock In {\em Proceedings of the IEEE conference on computer vision and
  pattern recognition}, pages 3213--3223, 2016.

\bibitem{de2017semantic}
Bert De~Brabandere, Davy Neven, and Luc Van~Gool.
\newblock Semantic instance segmentation with a discriminative loss function.
\newblock {\em arXiv preprint arXiv:1708.02551}, 2017.

\bibitem{degeus2019fast}
Daan {de Geus}, Panagiotis Meletis, and Gijs Dubbelman.
\newblock Fast panoptic segmentation network.
\newblock {\em arXiv preprint arXiv:1910.03892}, 2019.

\bibitem{degeus2019panoptic}
Daan {de Geus}, Panagiotis Meletis, and Gijs Dubbelman.
\newblock Single network panoptic segmentation for street scene understanding.
\newblock In {\em 2019 IEEE Intelligent Vehicles Symposium (IV)}, 2019.

\bibitem{deng2009imagenet}
Jia Deng, Wei Dong, Richard Socher, Li-Jia Li, Kai Li, and Li Fei-Fei.
\newblock Imagenet: A large-scale hierarchical image database.
\newblock In {\em 2009 IEEE conference on computer vision and pattern
  recognition}, pages 248--255. Ieee, 2009.

\bibitem{gao2019ssap}
Naiyu Gao, Yanhu Shan, Yupei Wang, Xin Zhao, Yinan Yu, Ming Yang, and Kaiqi
  Huang.
\newblock Ssap: Single-shot instance segmentation with affinity pyramid.
\newblock In {\em Proceedings of the IEEE International Conference on Computer
  Vision}, pages 642--651, 2019.

\bibitem{goodman2001classes}
Joshua Goodman.
\newblock Classes for fast maximum entropy training.
\newblock In {\em 26th International Conference on Acoustics, Speech, and
  Signal Processing}, 2001.

\bibitem{guerriero2018deep}
Samantha Guerriero, Barbara Caputo, and Thomas Mensink.
\newblock Deepncm: Deep nearest class mean classifiers.
\newblock In {\em 6th International Conference on Learning Representations,
  {ICLR} 2018, Vancouver, BC, Canada, April 30 - May 3, 2018, Workshop Track
  Proceedings}, 2018.

\bibitem{he2017mask}
Kaiming He, Georgia Gkioxari, Piotr Doll{\'a}r, and Ross Girshick.
\newblock Mask r-cnn.
\newblock In {\em Proceedings of the IEEE international conference on computer
  vision}, pages 2961--2969, 2017.

\bibitem{he2016deep}
Kaiming He, Xiangyu Zhang, Shaoqing Ren, and Jian Sun.
\newblock Deep residual learning for image recognition.
\newblock In {\em Proceedings of the IEEE conference on computer vision and
  pattern recognition}, pages 770--778, 2016.

\bibitem{hou2020real}
Rui Hou, Jie Li, Arjun Bhargava, Allan Raventos, Vitor Guizilini, Chao Fang,
  Jerome Lynch, and Adrien Gaidon.
\newblock Real-time panoptic segmentation from dense detections.
\newblock In {\em Proceedings of the IEEE/CVF Conference on Computer Vision and
  Pattern Recognition}, pages 8523--8532, 2020.

\bibitem{hwang2019segsort}
Jyh-Jing Hwang, Stella~X Yu, Jianbo Shi, Maxwell~D Collins, Tien-Ju Yang, Xiao
  Zhang, and Liang-Chieh Chen.
\newblock Segsort: Segmentation by discriminative sorting of segments.
\newblock In {\em Proceedings of the IEEE/CVF International Conference on
  Computer Vision}, pages 7334--7344, 2019.

\bibitem{kingma2014adam}
Diederik Kingma and Jimmy Ba.
\newblock Adam: A method for stochastic optimization.
\newblock {\em International Conference on Learning Representations}, 12 2014.

\bibitem{kirillov2019panopticfpn}
Alexander Kirillov, Ross Girshick, Kaiming He, and Piotr Doll{\'a}r.
\newblock Panoptic feature pyramid networks.
\newblock In {\em Proceedings of the IEEE Conference on Computer Vision and
  Pattern Recognition}, pages 6399--6408, 2019.

\bibitem{kirillov2019panoptic}
Alexander Kirillov, Kaiming He, Ross Girshick, Carsten Rother, and Piotr
  Doll{\'a}r.
\newblock Panoptic segmentation.
\newblock In {\em Proceedings of the IEEE Conference on Computer Vision and
  Pattern Recognition}, pages 9404--9413, 2019.

\bibitem{law2018cornernet}
Hei Law and Jia Deng.
\newblock Cornernet: Detecting objects as paired keypoints.
\newblock In {\em Proceedings of the European Conference on Computer Vision
  (ECCV)}, pages 734--750, 2018.

\bibitem{lazarow2019learning}
Justin Lazarow, Kwonjoon Lee, and Zhuowen Tu.
\newblock Learning instance occlusion for panoptic segmentation.
\newblock {\em arXiv preprint arXiv:1906.05896}, 2019.

\bibitem{li2018learning}
Jie Li, Allan Raventos, Arjun Bhargava, Takaaki Tagawa, and Adrien Gaidon.
\newblock Learning to fuse things and stuff.
\newblock {\em arXiv preprint arXiv:1812.01192}, 2018.

\bibitem{li2018weakly}
Qizhu Li, Anurag Arnab, and Philip~HS Torr.
\newblock Weakly-and semi-supervised panoptic segmentation.
\newblock In {\em Proceedings of the European Conference on Computer Vision
  (ECCV)}, pages 102--118, 2018.

\bibitem{li2019attention}
Yanwei Li, Xinze Chen, Zheng Zhu, Lingxi Xie, Guan Huang, Dalong Du, and
  Xingang Wang.
\newblock Attention-guided unified network for panoptic segmentation.
\newblock In {\em Proceedings of the IEEE Conference on Computer Vision and
  Pattern Recognition}, pages 7026--7035, 2019.

\bibitem{lin2014microsoft}
Tsung-Yi Lin, Michael Maire, Serge Belongie, James Hays, Pietro Perona, Deva
  Ramanan, Piotr Doll{\'a}r, and C~Lawrence Zitnick.
\newblock Microsoft coco: Common objects in context.
\newblock In {\em European conference on computer vision}, pages 740--755.
  Springer, 2014.

\bibitem{liu2019end}
Huanyu Liu, Chao Peng, Changqian Yu, Jingbo Wang, Xu Liu, Gang Yu, and Wei
  Jiang.
\newblock An end-to-end network for panoptic segmentation.
\newblock In {\em Proceedings of the IEEE Conference on Computer Vision and
  Pattern Recognition}, pages 6172--6181, 2019.

\bibitem{liu2018path}
Shu Liu, Lu Qi, Haifang Qin, Jianping Shi, and Jiaya Jia.
\newblock Path aggregation network for instance segmentation.
\newblock In {\em Proceedings of the IEEE Conference on Computer Vision and
  Pattern Recognition}, pages 8759--8768, 2018.

\bibitem{long2015fully}
Jonathan Long, Evan Shelhamer, and Trevor Darrell.
\newblock Fully convolutional networks for semantic segmentation.
\newblock In {\em Proceedings of the IEEE conference on computer vision and
  pattern recognition}, pages 3431--3440, 2015.

\bibitem{mikolov2013distributed}
Tomas Mikolov, Ilya Sutskever, Kai Chen, Greg~S Corrado, and Jeff Dean.
\newblock Distributed representations of words and phrases and their
  compositionality.
\newblock In {\em Advances in neural information processing systems}, pages
  3111--3119, 2013.

\bibitem{miller1995wordnet}
George~A Miller.
\newblock Wordnet: a lexical database for english.
\newblock {\em Communications of the ACM}, 38(11):39--41, 1995.

\bibitem{mnih2009scalable}
Andriy Mnih and Geoffrey~E Hinton.
\newblock A scalable hierarchical distributed language model.
\newblock In {\em Advances in neural information processing systems}, pages
  1081--1088, 2009.

\bibitem{morin2005hierarchical}
Frederic Morin and Yoshua Bengio.
\newblock Hierarchical probabilistic neural network language model.
\newblock In {\em Aistats}, volume~5, pages 246--252. Citeseer, 2005.

\bibitem{neuhold2017mapillary}
Gerhard Neuhold, Tobias Ollmann, Samuel Rota~Bul\`o, and Peter Kontschieder.
\newblock The mapillary vistas dataset for semantic understanding of street
  scenes.
\newblock In {\em International Conference on Computer Vision (ICCV)}, 2017.

\bibitem{neven2019instance}
Davy Neven, Bert~De Brabandere, Marc Proesmans, and Luc~Van Gool.
\newblock Instance segmentation by jointly optimizing spatial embeddings and
  clustering bandwidth.
\newblock In {\em Proceedings of the IEEE Conference on Computer Vision and
  Pattern Recognition}, pages 8837--8845, 2019.

\bibitem{neven2018towards}
Davy Neven, Bert De~Brabandere, Stamatios Georgoulis, Marc Proesmans, and Luc
  Van~Gool.
\newblock Towards end-to-end lane detection: an instance segmentation approach.
\newblock In {\em 2018 IEEE Intelligent Vehicles Symposium (IV)}, pages
  286--291. IEEE, 2018.

\bibitem{newell2017pixels}
Alejandro Newell and Jia Deng.
\newblock Pixels to graphs by associative embedding.
\newblock In {\em Advances in neural information processing systems}, pages
  2171--2180, 2017.

\bibitem{newell2017associative}
Alejandro Newell, Zhiao Huang, and Jia Deng.
\newblock Associative embedding: End-to-end learning for joint detection and
  grouping.
\newblock In {\em Advances in Neural Information Processing Systems}, pages
  2277--2287, 2017.

\bibitem{porzi2019seamless}
Lorenzo Porzi, Samuel~Rota Bulo, Aleksander Colovic, and Peter Kontschieder.
\newblock Seamless scene segmentation.
\newblock In {\em Proceedings of the IEEE Conference on Computer Vision and
  Pattern Recognition}, pages 8277--8286, 2019.

\bibitem{razavi2019generating}
Ali Razavi, Aaron van~den Oord, and Oriol Vinyals.
\newblock Generating diverse high-fidelity images with vq-vae-2.
\newblock {\em arXiv preprint arXiv:1906.00446}, 2019.

\bibitem{redmon2017yolo9000}
Joseph Redmon and Ali Farhadi.
\newblock Yolo9000: better, faster, stronger.
\newblock In {\em Proceedings of the IEEE conference on computer vision and
  pattern recognition}, pages 7263--7271, 2017.

\bibitem{sandler2018mobilenetv2}
Mark Sandler, Andrew Howard, Menglong Zhu, Andrey Zhmoginov, and Liang-Chieh
  Chen.
\newblock Mobilenetv2: Inverted residuals and linear bottlenecks.
\newblock In {\em Proceedings of the IEEE conference on computer vision and
  pattern recognition}, pages 4510--4520, 2018.

\bibitem{thomson1904structure}
JJ Thomson.
\newblock On the structure of the atom: an investigation of the stability and
  periods of oscillation of a number of corpuscles arranged at equal intervals
  around the circumference of a circle; with application of the results to the
  theory of atomic structure.
\newblock {\em Philos. Mag., Ser. 6}, 7:237--265, 1904.

\bibitem{tighe2014scene}
Joseph Tighe, Marc Niethammer, and Svetlana Lazebnik.
\newblock Scene parsing with object instances and occlusion ordering.
\newblock In {\em Proceedings of the IEEE Conference on Computer Vision and
  Pattern Recognition}, pages 3748--3755, 2014.

\bibitem{tu2005image}
Zhuowen Tu, Xiangrong Chen, Alan~L Yuille, and Song-Chun Zhu.
\newblock Image parsing: Unifying segmentation, detection, and recognition.
\newblock {\em International Journal of computer vision}, 63(2):113--140, 2005.

\bibitem{wang2020solo}
Xinlong Wang, Tao Kong, Chunhua Shen, Yuning Jiang, and Lei Li.
\newblock {SOLO}: Segmenting objects by locations.
\newblock In {\em Proc. Eur. Conf. Computer Vision (ECCV)}, 2020.

\bibitem{wang2019associatively}
Xinlong Wang, Shu Liu, Xiaoyong Shen, Chunhua Shen, and Jiaya Jia.
\newblock Associatively segmenting instances and semantics in point clouds.
\newblock In {\em Proceedings of the IEEE Conference on Computer Vision and
  Pattern Recognition}, pages 4096--4105, 2019.

\bibitem{xiong2019upsnet}
Yuwen Xiong, Renjie Liao, Hengshuang Zhao, Rui Hu, Min Bai, Ersin Yumer, and
  Raquel Urtasun.
\newblock Upsnet: A unified panoptic segmentation network.
\newblock In {\em Proceedings of the IEEE Conference on Computer Vision and
  Pattern Recognition}, pages 8818--8826, 2019.

\bibitem{yang2019deeperlab}
Tien-Ju Yang, Maxwell~D Collins, Yukun Zhu, Jyh-Jing Hwang, Ting Liu, Xiao
  Zhang, Vivienne Sze, George Papandreou, and Liang-Chieh Chen.
\newblock Deeperlab: Single-shot image parser.
\newblock {\em arXiv preprint arXiv:1902.05093}, 2019.

\bibitem{yang2020sognet}
Yibo Yang, Hongyang Li, Xia Li, Qijie Zhao, Jianlong Wu, and Zhouchen Lin.
\newblock Sognet: Scene overlap graph network for panoptic segmentation.
\newblock In {\em Proceedings of the AAAI Conference on Artificial
  Intelligence}, volume~34, pages 12637--12644, 2020.

\bibitem{yao2012describing}
Jian Yao, Sanja Fidler, and Raquel Urtasun.
\newblock Describing the scene as a whole: Joint object detection, scene
  classification and semantic segmentation.
\newblock In {\em 2012 {IEEE} Conference on Computer Vision and Pattern
  Recognition, Providence, RI, USA, June 16-21, 2012}, pages 702--709, 2012.

\bibitem{zhao2017pyramid}
Hengshuang Zhao, Jianping Shi, Xiaojuan Qi, Xiaogang Wang, and Jiaya Jia.
\newblock Pyramid scene parsing network.
\newblock In {\em Proceedings of the IEEE conference on computer vision and
  pattern recognition}, pages 2881--2890, 2017.

\bibitem{zhou2019objects}
Xingyi Zhou, Dequan Wang, and Philipp Kr{\"a}henb{\"u}hl.
\newblock Objects as points.
\newblock {\em arXiv preprint arXiv:1904.07850}, 2019.

\end{thebibliography}
}

{\ifsuppmat
\clearpage
{
\twocolumn[
  \begin{@twocolumnfalse}

   \newpage
   \null
   \iftoggle{cvprrebuttal}{\vspace*{-.3in}}{\vskip .375in}
   \begin{center}
      {\Large \bf \newtitle\\-- Supplementary Material -- \par}  %
      \iftoggle{cvprrebuttal}{\vspace*{-22pt}}{\vspace*{24pt}}
      {
      \large
      \lineskip .5em
      \begin{tabular}[t]{c}
        \iftoggle{cvprfinal}{
          \authorlist{}
        }{
          \iftoggle{cvprrebuttal}{}{
            Anonymous \confYear~submission\\
            \vspace*{1pt}\\
            Paper ID \cvprPaperID 
          }
        }
      \end{tabular}
      \par
      }
      \vskip .5em
      \vspace*{12pt}
   \end{center}\

  \end{@twocolumnfalse}
  ]

}
\setcounter{section}{1}
\setcounter{table}{0}
\setcounter{figure}{0}
\setcounter{equation}{0}
\renewcommand{\thesubsection}{\Alph{subsection}}
\renewcommand{\thetable}{\Alph{table}}
\renewcommand{\thefigure}{\Alph{figure}}
\renewcommand{\theequation}{\Alph{equation}}
\appendix

\section{Evaluation Metrics}
For completeness, and making the paper self-contained, we review the standard panoptic segmentation metrics used for our experiments in this section.

We evaluate our method using the standard metric panoptic quality (PQ)~\cite{kirillov2019panoptic} as well as its variations, $PQ^\dagger$~\cite{porzi2019seamless} and parsing covering (PC)~\cite{yang2019deeperlab}.
$PQ$ formulates the quality of the predicted panoptic segmentation in terms of
intersection over union (IoU), true positives (TP), false positives (FP) and false negatives (FN).
\begin{align}
    PQ = \frac{\sum_{(p, q) \in \texttt{TP}} \IoU(p, q) \mathds{1}_{\IoU(p, q) > 0.5}}{|\texttt{TP}| + \frac12|\texttt{FP}| + \frac12|\texttt{FN}|}~,
\end{align}%
\noindent where $(p, q)$ is the tuple of the predicted and ground-truth mask, respectively.
Additionally, thing- and stuff-specific $PQ$ are denoted as $PQ_{th}$ and $PQ_{st}$. 
While popular in the literature, the $PQ$ metric has two downsides that has been pointed out in previous work~\cite{porzi2019seamless,yang2019deeperlab}.

First, the $PQ$ metric is harsh towards stuff classes and requires $IoU$ overlap larger than 0.5 even for stuff, treating it like an instance.
The $PQ^\dagger$ metric~\cite{porzi2019seamless} aims to mitigate this by relaxing the IoU threshold to 0 for stuff classes, and calculates the $PQ$ metric as usual for thing classes.
Second, the $PQ$ metric treats objects the same regardless of size, making it very sensitive to small false positives.

The parsing covering (PC)~\cite{yang2019deeperlab} metric targets applications where larger objects are more important, such as autonomous driving, and is defined as
\begin{align}
    PC = \frac{1}{|\set{C}|} \sum_{c \in \set{C}} \frac{\sum_{R \in \set{R}_c} |R| \max_{R' \in {\set{R}'}_c} \IoU(R', R)}{\sum_{R \in \set{R}_c} |R|}~,
\end{align}%
\noindent where $\set{R}_c$, ${\set{R}'}_c$ are the ground-truth and predicted regions of class $c$, respectively.

\section{Additional Metrics Experimental Results}
In this section, we describe the experimental results on Cityscapes with the alternative panoptic segmentation metrics $PQ^\dagger$~\cite{porzi2019seamless} and parsing covering (PC)~\cite{yang2019deeperlab},
as well as the related tasks semantic segmentation and instance segmentation metrics $mIoU$ and $AP$. We include $PQ_{th}$ to facilitate comparison with $AP$.

The results for $PQ^\dagger$ and $PC$ can be seen in Tables~\ref{tbl:cityscapes-val-pq-plus} and~\ref{tbl:cityscapes-val-pc}.
We can see that our proposed method is able to get competitive results in terms of $PQ^\dagger$ and $PC$, even when comparing with proposal-based methods.
Particularly, we note that our method outperforms the method of Porzi et al.~\cite{porzi2019seamless} in terms of the $PQ^\dagger$ metric, indicating that our model is handling stuff classes well, which is also illustrated by our model outperforming all others in terms of the $PQ_{st}$ stuff metric.
Our method being competitive in terms of the $PC$ metric indicates that large objects are segmented well.

In Table~\ref{tbl:cityscapes-val-miou-ap}, we report the sub-task metrics $mIo$U and $AP$ for reference. We noticed that although our method achieves similar $PQ_{th}$ with the others, the gap in $AP$ is relatively apparent. The difference between $PQ$ and $AP$ in evaluating instance segmentation performance lies in how the acceptance threshold for objectness score (or detection confidence) is handled.
Unlike $PQ$, which uses a fixed threshold, the $AP$ metric relies heavily on the score estimation of each instance mask to be able to estimate the optimal threshold during evaluation.
Notably, $PQ$ is a quite different metric from $AP$, where false positives matter a lot.
Consider the definition of the $AP$ metric:
\begin{align}
    AP = \frac{1}{|\set{R}|} \sum_{r \in \set{R}} \max_{\hat{r} \ge r} P(\hat{r})~,
\end{align}
\noindent where $P(r)$ is the precision at recall $r$ and $\set{R}$ is the set of recall levels.
The $AP$ evaluation protocol uses all possible thresholds that provide the requested recall levels in order to evaluate the model, while $PQ$ evaluation requires finding a single threshold (\eg $r=r_0$) that works for all images in the dataset. Arguably, it can be said that $PQ$ evaluation is closer to representing the performance of the model in a real production environment, where a single fixed threshold must be set for inference.
It would be an interesting future direction to explore how we can improve $AP$ at the same time under our algorithm paradigm.

\section{Alternative Postprocessing Algorithm with Downsampling}
In this section, we discuss an alternative postprocessing algorithm which increases speed at small cost of $PQ$.
It is possible to speed up the postprocessing of our method by operating on a downsampled version of the embedding space. The resulting postprocessing time and how it affects Cityscapes validation set $PQ$ can be seen in Figure~\ref{fig:postprocessing-speed}. The downsampling factor refers to how much smaller the spatial size of the embedding space we operate postprocessing on becomes. For example, a $1024 \times 2048$ size embedding space with downsampling factor $4$ becomes $256 \times 512$, reducing postprocessing time to $8$ ms, while only reducing Cityscapes validation set $PQ$ to $58.4$. This simple modification of the postprocessing algorithm can increase inference speed at slight cost of accuracy.
Therefore, it can be decided whether to weigh accuracy or speed higher, or have a mixture of both, with this simple modification.
\vspace{10em}%

\begin{figure}[t]
{\centering
\def\figwidth{1.0}
\includegraphics[width=\figwidth\linewidth]{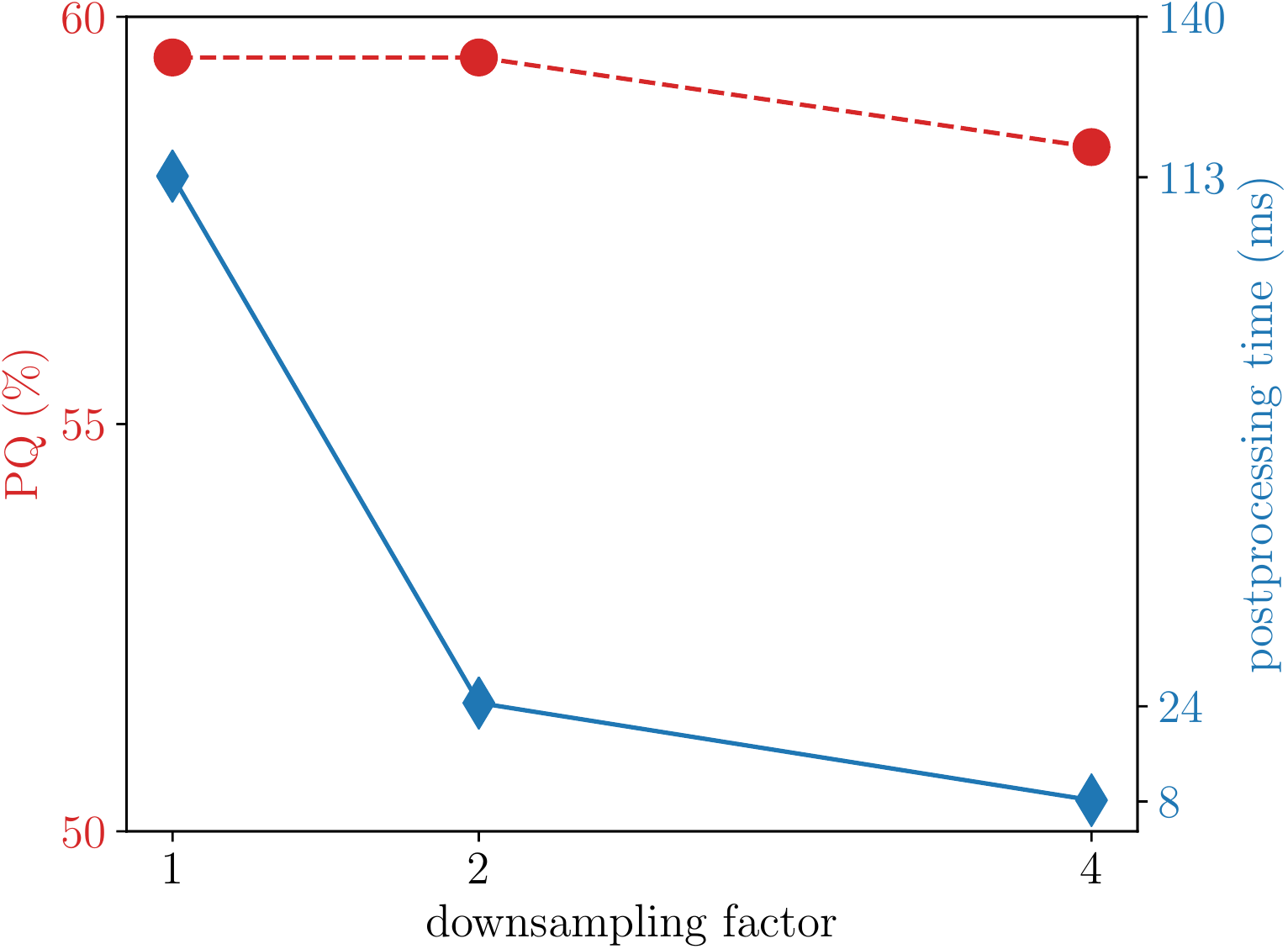}
   \caption{Postprocessing time (solid line) vs. $PQ$ (dashed line) as a function of the downsampling factor used in the postprocessing.}%
   \label{fig:postprocessing-speed}%
}%
\end{figure}%

\begin{table}[th]
\centering
\begin{tabular}{|l|l|l|c|}
\hline
Method & Backbone & Pretrain. & $PQ^\dagger$ \\
\hline
\multicolumn{4}{l}{Proposal-based}  \\
\hline
Seamless~\cite{porzi2019seamless} & ResNet50 & ImageNet & $59.6$ \\
\hline
\multicolumn{4}{l}{Proposal-free} \\
\hline
\textbf{HLE (Ours)}& ResNet50 & ImageNet & $\mathbf{61.3}$ \\
\hline
\end{tabular}
\vspace{0.5em}
\caption{Single-scale experimental results on the Cityscapes validation set.}
\label{tbl:cityscapes-val-pq-plus}
\end{table}%
\setlength{\tabcolsep}{1.4pt}
\begin{table}[th]
\centering
\begin{tabular}{|l|l|l|c|}
\hline
Method & Backbone & Pretrain. & $PC$ \\
\hline
\multicolumn{4}{l}{Proposal-free} \\
\hline
DeeperLab~\cite{yang2019deeperlab} & Xception71 & ImageNet & $75.6$ \\
DeeperLab~\cite{yang2019deeperlab} & Wider MNV2 & ImageNet & $74.0$ \\
DeeperLab~\cite{yang2019deeperlab} & L. W. MNV2 & ImageNet & $67.9$ \\
\hline
\textbf{HLE (Ours)}& ResNet50 & ImageNet & $\mathbf{76.6}$ \\
\hline
\end{tabular}
\vspace{0.5em}
\caption{Single-scale experimental results on the Cityscapes validation set.}
\label{tbl:cityscapes-val-pc}
\end{table}%
\setlength{\tabcolsep}{1.4pt}
\setlength{\tabcolsep}{1.4pt}
\begin{table}[th]
\centering
\begin{tabular}{|l|l|l|c|c|c|}
\hline
Method & Backbone & Pretrain. & $mIoU$ & $AP$ & $PQ_{th}$ \\
\hline
\multicolumn{6}{l}{Proposal-based}  \\
\hline
Seamless~\cite{porzi2019seamless} & ResNet50 & ImageNet & $\mathbf{77.5}$ & 33.6 & $\mathbf{56.1}$ \\
Real-time PS~\cite{hou2020real} & ResNet50 & ImageNet & 77.0 & 29.8 & 52.1 \\
UPSNet~\cite{xiong2019upsnet} & ResNet50 & ImageNet &  75.2 & 33.3 & 54.6 \\  %
Pan. FPN~\cite{kirillov2019panopticfpn} & ResNet50 & ImageNet & 75.0 & 32.0 & 51.6 \\
Attn.-Guid.~\cite{li2019attention} & ResNet50 & ImageNet & 73.6 & 33.6 & 52.7 \\
Li et al.~\cite{li2018weakly} & ResNet101 & ImageNet & 71.6 & 24.3 & 39.6 \\
PANet~\cite{liu2018path} & ResNet50 & ImageNet & - & $\mathbf{36.5}$ & - \\
\hline
\multicolumn{6}{l}{Proposal-free} \\
\hline
SSAP~\cite{gao2019ssap} & ResNet50 & ImageNet & - & $\mathbf{32.8}$ & - \\
\hline
\textbf{HLE (Ours)}& ResNet50 & ImageNet & $\mathbf{77.3}$ & $23.9$ & $\mathbf{51.1}$\\
\hline
\end{tabular}
\vspace{0.5em}
\caption{Single-scale experimental results on the Cityscapes validation set.
}
\label{tbl:cityscapes-val-miou-ap}
\end{table}%
\setlength{\tabcolsep}{1.4pt}

}

\end{document}